\documentclass{article}
\usepackage{graphicx}
\usepackage{amsmath}
\usepackage{fontawesome5}
\usepackage{geometry}
\usepackage{csquotes}
\usepackage{comment}
\usepackage[table,xcdraw]{xcolor}
\usepackage{listings}
\usepackage{authblk}
\usepackage{lipsum}
\usepackage{minitoc}
\usepackage{booktabs} 
\usepackage{longtable}
\usepackage{array}
\usepackage[justification=justified]{caption}
\usepackage{lmodern}
\usepackage{orcidlink}
\usepackage{placeins}

\hypersetup{
    colorlinks=true,
    linkcolor=[RGB]{0,51,153},  
    urlcolor=[RGB]{0,51,153}    
    citecolor=[RGB]{0,51,153},    
    filecolor=[RGB]{0,51,153},
    allcolors=[RGB]{0,51,153} 
}

\noptcrule 
\mtcsetfont{parttoc}{section}{\fontsize{7.5pt}{7.5pt}\sffamily}

\newcommand{\github}[1]{%
   \href{#1}{\faGithubSquare}%
}

\title{34 Examples of LLM Applications in Materials Science and Chemistry: Towards Automation, Assistants, Agents, and Accelerated Scientific Discovery}

\author[1]{Yoel Zimmermann\orcidlink{0009-0003-1720-4368}}
\author[2]{Adib Bazgir\orcidlink{0000-0001-6475-8505}}
\author[3]{Alexander Al-Feghali\orcidlink{0009-0004-8377-7049}}
\author[4]{Mehrad Ansari\orcidlink{0000-0001-5696-9193}}
\author[27]{Joshua Bocarsly\orcidlink{0000-0002-7523-152X}}
\author[5]{L. Catherine Brinson\orcidlink{0000-0003-2551-1563}}
\author[6,7]{Yuan Chiang\orcidlink{0000-0002-4017-7084}}
\author[5]{Defne Circi\orcidlink{0000-0002-5761-0198}}
\author[8]{Min-Hsueh Chiu\orcidlink{0000-0003-0637-7856}}
\author[9]{Nathan Daelman\orcidlink{0000-0002-7647-1816}}
\author[10,11]{Matthew L. Evans\orcidlink{0000-0002-1182-9098}}
\author[26]{Abhijeet S. Gangan}
\author[12,13]{Janine George\orcidlink{0000-0001-8907-0336}} 
\author[14]{Hassan Harb\orcidlink{0000-0002-6016-3122}}
\author[5]{Ghazal Khalighinejad\orcidlink{0009-0005-2476-8043}}
\author[15]{Sartaaj Takrim Khan\orcidlink{0009-0009-2131-9700}}
\author[9]{Sascha Klawohn\orcidlink{0000-0003-4850-776X}}
\author[1,20]{Magdalena Lederbauer\orcidlink{0009-0008-0665-1839}}
\author[16]{Soroush Mahjoubi\orcidlink{0000-0001-8879-5431}}
\author[9,17]{Bernadette Mohr\orcidlink{0000-0003-0903-0073}}
\author[4,15]{Seyed Mohamad Moosavi\orcidlink{0000-0002-0357-5729}}
\author[12,13]{Aakash Naik\orcidlink{0000-0002-6071-6786}}
\author[16]{Aleyna Beste Ozhan\orcidlink{0000-0002-0281-3860}}
\author[18]{Dieter Plessers\orcidlink{0000-0001-8906-8447}}
\author[19]{Aritra Roy\orcidlink{0000-0003-0243-9124}}
\author[9]{Fabian Schöppach}
\author[20]{Philippe Schwaller\orcidlink{0000-0003-3046-6576}}
\author[21]{Carla Terboven}
\author[12,13]{Katharina Ueltzen\orcidlink{0009-0003-2967-1182}}
\author[28]{Yue Wu\orcidlink{0000-0003-2874-8267}}
\author[22]{Shang Zhu\orcidlink{0000-0002-8433-8599}}
\author[23]{Jan Janssen\orcidlink{0000-0001-9948-7119}}
\author[24]{Calvin Li}
\author[14,25]{Ian Foster\orcidlink{0000-0003-2129-5269}}
\author[14,25]{Ben Blaiszik$^*$\orcidlink{0000-0002-5326-4902}} 

\affil[1]{ETH Zurich}
\affil[2]{University of Missouri-Columbia}
\affil[3]{McGill University}
\affil[4]{Acceleration Consortium}
\affil[5]{Duke University}
\affil[6]{University of California at Berkeley}
\affil[7]{Lawrence Berkeley National Laboratory}
\affil[8]{University of Southern California}
\affil[9]{Humboldt University of Berlin}
\affil[10]{Université catholique de Louvain}
\affil[11]{Matgenix SRL}
\affil[12]{Friedrich-Schiller-Universität Jena}
\affil[13]{Federal Institute of Materials Research and Testing (BAM)}
\affil[14]{Argonne National Laboratory}
\affil[15]{University of Toronto}
\affil[16]{Massachusetts Institute of Technology}
\affil[17]{University of Amsterdam}
\affil[18]{KU Leuven}
\affil[19]{London South Bank University}
\affil[20]{EPFL}
\affil[21]{Helmholtz-Zentrum Berlin für Materialien und Energie GmbH}
\affil[22]{University of Michigan-Ann Arbor}
\affil[23]{Max-Planck Institute for Sustainable Materials}
\affil[24]{Fum Technologies, Inc.}
\affil[25]{University of Chicago}
\affil[26]{University of California, Los Angeles}
\affil[27]{University of Houston}
\affil[28]{Independent Researcher}

\date{}

\begin{document}
\doparttoc 
\faketableofcontents 
\maketitle
\thanks{$^*$Corresponding author: blaiszik@uchicago.edu}

\begin{abstract}
    Large Language Models (LLMs) are reshaping many aspects of materials science and chemistry research, enabling advances in molecular property prediction, materials design, scientific automation, knowledge extraction, and more. Recent developments demonstrate that the latest class of models are able to integrate structured and unstructured data, assist in hypothesis generation, and streamline research workflows. To explore the frontier of LLM capabilities across the research lifecycle, we review applications of LLMs through 34 total projects developed during the second annual Large Language Model Hackathon for Applications in Materials Science and Chemistry, a global hybrid event. These projects spanned seven key research areas: (1) molecular and material property prediction, (2) molecular and material design, (3) automation and novel interfaces, (4) scientific communication and education, (5) research data management and automation, (6) hypothesis generation and evaluation, and (7) knowledge extraction and reasoning from the scientific literature. Collectively, these applications illustrate how LLMs serve as versatile predictive models, platforms for rapid prototyping of domain-specific tools, and much more. In particular, improvements in both open source and proprietary LLM performance through the addition of reasoning, additional training data, and new techniques have expanded effectiveness, particularly in low-data environments and interdisciplinary research. As LLMs continue to improve, their integration into scientific workflows presents both new opportunities and new challenges, requiring ongoing exploration, continued refinement, and further research to address reliability, interpretability, and reproducibility. 
\end{abstract}

\section*{Introduction}
\begin{figure}
    \centering
    \includegraphics[width=0.85\linewidth]{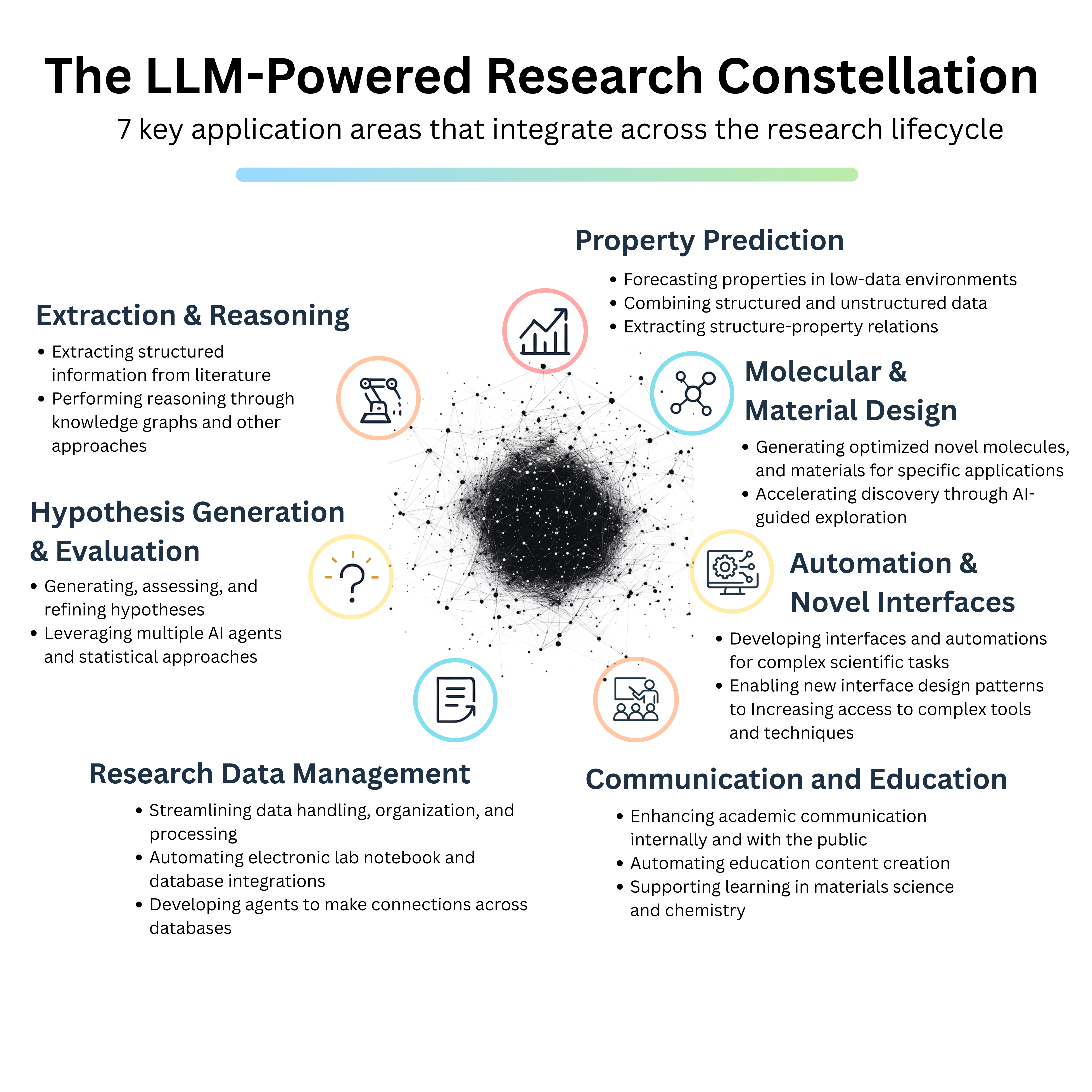}
    \caption{The LLM-Powered Research Constellation. At each stage of the research process, from initial ideation through experimental execution and communication of results, LLMs provide a constellation of capabilities spanning hypothesis generation, property prediction, novel interfaces, education, material design, automation, data management, scientific communication, and more. This constellation demonstrates the possibility of LLMs and multimodal models to drive a more efficient, rapid, and creative scientific discovery process through integrations across the research lifecycle.}
    \label{fig:overview}
\end{figure}

The integration of large language models (LLMs) into scientific workflows is reshaping how researchers approach data-driven discovery, automation, and even scientific reasoning and hypothesis generation~\cite{Jablonka_Schwaller_Ortega-Guerrero_Smit_2024,Bhattacharya2024Large, choudhary2024atomgpt, yin2024comparative}. In chemistry and materials science, fields characterized by complex data modalities, heterogeneous data formats, sparse experimental datasets, and fragmented knowledge ecosystems, LLMs are emerging as versatile tools capable of bridging gaps between computational methods, experimental data, literature and text sources, and domain expertise~\cite{gupta2022matscibert,liu2025alchembert,rubungo2023llm,kim2024explainable,ganose2019robocrystallographer,dunn2020benchmarking,petretto2018high, BazgirAdibICLR}.  Early applications have already demonstrated potential applicability in tasks ranging from molecular property prediction~\cite{jacobs2024regressionlargelanguagemodels, rubungo2024llm4mat, qian2023largelanguagemodelsempower} to automated laboratory workflows~\cite{Boiko2023Autonomous, Tom2024Self} and development of novel user interfaces~\cite{Zhang2024HoneyComb, Bran2023ChemCrow}. As illustrated in \autoref{fig:overview}, we note that there is a significant opportunity for these broad new capabilities to be incorporated throughout the scientific research lifecycle; from initial ideation through experimental execution to communication, learning, and further iteration.

However, the rapidity of change and the nearly constant release of models with higher performance, lower cost, and wider application spaces, and release of other platform capabilities (e.g., agentic tools, deep research modalities) make it challenging to keep pace, necessitating a collaborative and interdisciplinary effort to identify high-impact use cases, address specific limitations, and prototype applications to catalyze deeper study~\cite{Dubey2024Llama,Abdin2024Phi,Claude3,Jiang2024Mixtral,GPT4Turbo,GPT35Turbo,OpenAI2023GPT}. Towards this goal, we believe that accessing the wisdom of the crowd via science hackathons provides a powerful, and dynamic framework for fostering collaboration building, knowledge exchange, innovation, and incentivizing the rapid problem-solving and exploration needed to realize the benefit of these new models for scientific discovery in materials science and chemistry~\cite{nolte2020support,pe2019understanding,heller2023hack,hackathon-2023}.


In this work, we describe and analyze select applications developed as part of the second Large Language Model Hackathon for Applications in Materials Science and Chemistry~\cite{hackathon-2023}, detailing the broad classes of problems addressed by teams and highlighting trends in the approaches taken. We categorize the 34 submissions into seven key research areas and provide an overview of team contributions with highlights drawn from exemplar projects in each research area. We also present a summary table containing team details and code repository links for all submissions to offer a comprehensive view of the innovations demonstrated during the event.

Finally, we discuss the broader conclusions of the hackathon, emphasizing its role in fostering interdisciplinary collaboration, accelerating the adoption of artificial intelligence (AI) in scientific research~\cite{nolte2020support, pe2019understanding, heller2023hack}, and identifying key challenges that require further investigation. By examining these contributions, we provide insight into how structured collaborative frameworks can drive the systematic integration of LLMs into chemistry and materials science to accelerate research, improve researcher efficiency, and shape the future of AI-driven discovery.

\section*{Overview of Submissions}

The hackathon resulted in 34 team submissions (with 32 submissions providing detailed descriptions), covering a broad spectrum of materials science and chemistry applications. The submissions and links to the respective source code repositories are listed in \autoref{tab:submissions}. We categorized projects based on their primary objectives, clustering them into seven key areas, forming a constellation of new capabilities across the research lifecycle:

\begin{enumerate}
    \item \textbf{Molecular and Material Property Prediction}: Forecasting chemical and physical properties of molecules and materials using LLMs, particularly excelling in low-data environments and combining structured and unstructured data.

   \item \textbf{Molecular and Material Design}: Generating and optimizing novel molecules and materials using LLMs, including peptides, metal-organic frameworks, and sustainable construction materials.

    \item \textbf{Automation and Novel Interfaces}: Developing natural language interfaces and LLM-powered automated workflows to simplify complex scientific tasks, making advanced tools and techniques more accessible to researchers.

    \item \textbf{Scientific Communication and Education}: Enhancing academic communication, automating educational content creation, and supporting  learning in materials science and chemistry.

    \item \textbf{Research Data Management and Automation}: Streamlining the handling, organization, and processing of scientific data through LLM-powered tools and multimodal agents.

    \item \textbf{Hypothesis Generation and Evaluation}: Using LLMs to generate, assess, and refine scientific hypotheses, leveraging multiple AI agents and statistical approaches.

    \item \textbf{Knowledge Extraction and Reasoning}: Extracting structured information from scientific literature and performing sophisticated reasoning about chemical and materials science concepts through knowledge graphs and multimodal approaches.
    
\end{enumerate}
Collectively, this constellation of capabilities, shown in ~\autoref{fig:overview}, is applicable to long-standing challenges across the research lifecycle, creating a flywheel of improvements that promises to empower researchers with new capabilities and to speed the research process.
\renewcommand{\arraystretch}{1.5} 

\begin{longtable}{>{\raggedright\arraybackslash}p{0.50\linewidth} >{\raggedright\arraybackslash}p{0.35\linewidth} >{\raggedright\arraybackslash}p{0.08\linewidth}}
\caption{Overview of the tools developed by the various tools, and links to source code repositories. Full descriptions of the projects can be found in Ref. \cite{zimmermann2025reflections2024largelanguage}.\label{tab:submissions}}\\
\toprule
\textbf{Project} & \textbf{Authors} & \textbf{Links} \\
\midrule
\endfirsthead

\toprule
\textbf{Project} & \textbf{Authors} & \textbf{Links} \\
\midrule
\endhead

\bottomrule
\endfoot

\textbf{Molecular and Material Property Prediction}\\

Leveraging Orbital-Based Bonding Analysis Information in LLMs & Katharina Ueltzen, Aakash Naik, Janine George & \href{https://github.com/kaueltzen/LLM_Hackathon_2024}{GitHub} \\

Context-Enhanced Material Property Prediction (CEMPP) & Federico Ottomano, Elena Patyukova, Judith Clymo, Dmytro Antypov, Chi Zhang, Aritra Roy, Piyush Ranjan Maharana, Weijie Zhang, Xuefeng Liu, Erik Bitzek & \href{https://github.com/fedeotto/lpoolmat_llms}{GitHub} \\

MolFoundation: Benchmarking Chemistry LLMs on Predictive Tasks & Hassan Harb, Xuefeng Liu, Anastasiia Tsymbal, Oleksandr Narykov, Dana O'Connor, Shagun Maheshwari, Stanley Lo, Archit Vasan, Zartashia Afzal, Kevin Shen & \href{https://github.com/shagunm1210/MolFoundation}{GitHub} \\

3D Molecular Feature Vectors for Large Language Models & Jan Weinreich, Ankur K. Gupta, Amirhossein D. Naghdi, Alishba Imran&  \href{https://github.com/janweinreich/geometric-geniuses}{GitHub}\\

LLMSpectrometry & Tyler Josephson, Fariha Agbere, Kevin Ishimwe, Colin Jones, Charishma Puli, Samiha Sharlin, Hao Liu & \href{https://github.com/ATOMSLab/LLMSpectroscopy}{GitHub} \\

\midrule

\textbf{Molecular and Material Design}\\

MC-Peptide: An Agentic Workflow for Data-Driven Design of Macrocyclic Peptides & Andres M. Bran, Anna Borisova, Marcel M. Calderon, Mark Tropin, Rob Mills, Philippe Schwaller & \href{https://github.com/doncamilom/mc-peptide}{GitHub} \\

Leveraging AI Agents for Designing Low Band Gap Metal-Organic Frameworks & Mehrad Ansari, Sartaaj Takrim Khan, Mahyar Rajabi, Seyed Mohamad Moosavi, Amro Aswad &  \href{https://github.com/mehradans92/PoreVoyant}{GitHub}\\

How Low Can You Go? Leveraging Small LLMs for Material Design & Alessandro Canalicchio, Alexander Moßhammer, Tehseen Rug, Christoph Völker&  \href{https://github.com/sandrocan/LLMs-for-design-of-alkali-activated-concrete-formulations}{GitHub}\\
\midrule

\textbf{Automation and Novel Interfaces}\\

LangSim & Yuan Chiang, Giuseppe Fisicaro, Greg Juhasz, Sarom Leang, Bernadette Mohr, Utkarsh Pratiush, Francesco Ricci, Leopold Talirz, Pablo A. Unzueta, Trung Vo, Gabriel Vogel, Sebastian Pagel, Jan Janssen & \href{https://github.com/jan-janssen/LangSim}{GitHub} \\

LLMicroscopilot: assisting microscope operations through LLMs & Marcel Schloz, Jose C. Gonzalez & \href{https://gitlab.com/Schlozma/llm_autotem}{GitHub} \\

T2Dllama: Harnessing Language Model for Density Functional Theory (DFT) Parameter Suggestion & Chiku Parida, Martin H. Petersen & \href{https://github.com/chiku-parida/T2Dllama}{GitHub} \\

Materials Agent: An LLM-Based Agent with Tool-Calling Capabilities for Cheminformatics & Archit Datar, Kedar Dabhadkar & \href{https://github.com/dkedar7/materials-agent}{GitHub} \\

LLM with Molecular Augmented Token & Luis Pinto, Xuan Vu Nguyen, Tirtha Vinchurkar, Pradip Si, Suneel Kuman & \href{https://github.com/luispintoc/LLM-mol-encoder}{GitHub} \\
\midrule

\textbf{Scientific Communication and Education}\\

MaSTeA: Materials Science Teaching Assistant & Defne Circi, Abhijeet S. Gangan, Mohd Zaki & \href{https://github.com/abhijeetgangan/MaSTeA}{GitHub} \\

LLMy-Way & Ruijie Zhu, Faradawn Yang, Andrew Qin, Suraj Sudhakar, Jaehee Park, Victor Chen & \href{https://github.com/Ray16/LLM_My_way}{GitHub} \\

WaterLLM: Creating a Custom ChatGPT for Water Purification Using PromptEngineering Techniques & Viktoriia Baibakova, Maryam G. Fard, Teslim Olayiwola, Olga Taran & \href{https://github.com/ViktoriiaBaib/WaterLLM}{GitHub} \\
\midrule

\textbf{Research Data Management and Automation}\\

yeLLowhaMMer: A Multi-modal Tool-calling Agent for Accelerated Research Data Management & Matthew L. Evans, Benjamin Charmes, Vraj Patel, Joshua D. Bocarsly& \href{https://github.com/datalab-org/yellowhammer}{GitHub} \\

LLMads & Sarthak Kapoor, José M. Pizarro, Ahmed Ilyas, Alvin N. Ladines, Vikrant Chaudhary & \href{https://github.com/ka-sarthak/llmads/tree/llm-hackathon-submission}{GitHub} \\

NOMAD Query Reporter: Automating Research Data Narratives & Nathan Daelman, Fabian Schöppach, Carla Terboven, Sascha Klawohn, Bernadette Mohr & \href{https://github.com/ndaelman-hu/nomad_query_reporter}{GitHub} \\

Speech-schema-filling: Creating Structured Data Directly from Speech & Hampus Näsström, Julia Schumann, Michael Götte, José A. Márquez & \href{https://github.com/hampusnasstrom/speech-schema-filling}{GitHub} \\
\midrule

\textbf{Hypothesis Generation and Evaluation}\\

Leveraging LLMs for Bayesian Temporal Evaluation of Scientific Hypotheses & Marcus Schwarting & \href{https://github.com/meschw04/lk99_inference}{GitHub} \\

Multi-Agent Hypothesis Generation and Verification through Tree of Thoughts and Retrieval Augmented Generation & Aleyna Beste Ozhan, Soroush Mahjoubi & \href{https://github.com/soroushmahj/ThoughtfulBeavers}{GitHub} \\

ActiveScience & Min-Hsueh Chiu & \href{https://github.com/minhsueh/ActiveScience}{GitHub} \\

G-Peer-T: LLM Probabilities For Assessing Scientific Novelty and Nonsense & Alexander Al-Feghali, Sylvester Zhang & \href{https://github.com/alxfgh/G-Peer-T/}{GitHub} \\
\midrule

\textbf{Knowledge Extraction and Reasoning}\\

ChemQA &  Ghazal Khalighinejad, Shang Zhu, Xuefeng Liu & \href{https://github.com/ghazalkhalighinejad/multimodalQA}{GitHub} \\

LithiumMind - Leveraging Language Models for Understanding Battery Performance & Xinyi Ni, Zizhang Chen, Rongda Kang, Sheng-Lun Liao, Pengyu Hong, Sandeep Madireddy & \href{https://github.com/KKbeckang/LiGPT-Beta}{GitHub} \\

KnowMat: Transforming Unstructured Material Science Literature into Structured Knowledge & Hasan M. Sayeed, Ramsey Issa, Trupti Mohanty, Taylor Sparks & \href{https://github.com/sparks-sayeed/LLMs_for_Materials_and_Chemistry_Hackathon}{GitHub} \\

Ontosynthesis & Qianxiang Ai, Jiaru Bai, Kevin Shen, Jennifer D'Souza, Elliot Risch & \href{https://github.com/qai222/ontosynthesis}{GitHub} \\

Knowledge Graph RAG for Polymer Simulation & Jiale Shi, Weijie Zhang, Dandan Tang, Chi Zhang & \href{https://github.com/shijiale0609/KG-RAG-LLM-Polymers}{GitHub} \\

Synthetic Data Generation and Insightful Machine Learning for High Entropy Alloy Hydrides & Tapashree Pradhan, Devi Dutta Biswajeet & \href{https://github.com/tapashreepradhan/LLM-materialsChem-hack24}{GitHub} \\

Chemsense: Are large language models aligned with human chemical preference? &  Martiño Ríos-García, Nawaf Alampara, Mara Schilling-Wilhelmi, Abdelrahman Ibrahim, Kevin Maik Jablonka & \href{https://github.com/lamalab-org/chemsense}{GitHub}\\

GlossaGen & Magdalena Lederbauer, Dieter Plessers, Philippe Schwaller & \href{https://github.com/mlederbauer/glossagen}{GitHub} \\
\end{longtable}

We next discuss the constellation of capabilities in more detail and highlight exemplar projects across each key application~area.

\section{Molecular and Material Property Prediction}
LLMs have rapidly advanced in 
molecular and material property prediction, 
employing both textual and numerical data to forecast a wide range of properties. Recent studies~\cite{Jablonka_Schwaller_Ortega-Guerrero_Smit_2024, qian2023largelanguagemodelsempower, jacobs2024regressionlargelanguagemodels, choudhary2024atomgpt} show LLMs performing comparably to, or even surpassing, conventional machine learning methods, particularly in low-data environments. The flexibility in processing both structured and unstructured data~\cite{Brown2020Language}, as well as their general applicability to regression tasks~\cite{vacareanu2024from}, make LLMs a powerful tool for diverse predictive tasks in molecular and materials science. 

\subsection{Leveraging orbital-based bonding analysis information in LLMs for material property predictions}
Previous studies have used different strategies to learn material properties using LLMs, such as enriching graph neural network (GNN) features with LLM embeddings~\cite{yin2024comparative}, training domain-specific LLMs and customizing model architectures~\cite{gupta2022matscibert,liu2025alchembert,rubungo2023llm}, or fine-tuning general-purpose LLMs~\cite{kim2024explainable,rubungo2024llm4mat}. While exact strategies have differed, existing models predominantly operate on string representations of crystal structures primarily consisting of compositional and structural information commonly found in crystallographic information files (CIFs). Multiple studies have successfully utilized the text descriptions of structures~\cite{rubungo2023llm,rubungo2024llm4mat,kim2024explainable} that can be generated using the Robocrystallographer package~\cite{ganose2019robocrystallographer}. These descriptions consist of structural features like bond lengths, coordination polyhedra, lattice parameters, coordinates, structure type, and other descriptors. Other studies explored different string representations of compositional and structural information~\cite{yin2024comparative,liu2025alchembert,rubungo2024llm4mat}.

The team behind this submission emphasizes that, to their knowledge, no previous studies investigated including orbital-based bonding analysis information in LLMs for materials property prediction tasks. Thus, in this pilot study, the team tested including such descriptions in LLMs to predict the highest-frequency peak in their phonon density of states (DOS)~\cite{petretto2018high,dunn2020benchmarking}. This target is relevant to the thermal properties of materials and it is a tracked component of the MatBench benchmark project. A key hypothesis is that the inclusion of the bonding analysis information for this vibrational property will improve the LLM's performance, as previous studies demonstrated the importance of such bonding information for the same target via a Random Forest model~\cite{naik2023quantum}. 

To test this hypothesis, the team fine-tuned multiple Llama 3 models on the textual description of 1264 crystal structures in the benchmark dataset. The text descriptions were generated using two packages: the Robocrystallographer and LobsterPy package~\cite{naik2024lobsterpy}.  The text descriptions from Lobsterpy consist of orbital-based bonding analyses containing information on covalent bond strengths and antibonding states. The data used here is available on Zenodo~\cite{naik2023zenodo} and was generated as part of an earlier dataset publication~\cite{naik2023quantum}.

\begin{figure}
    \centering
    \includegraphics[width=1\linewidth]{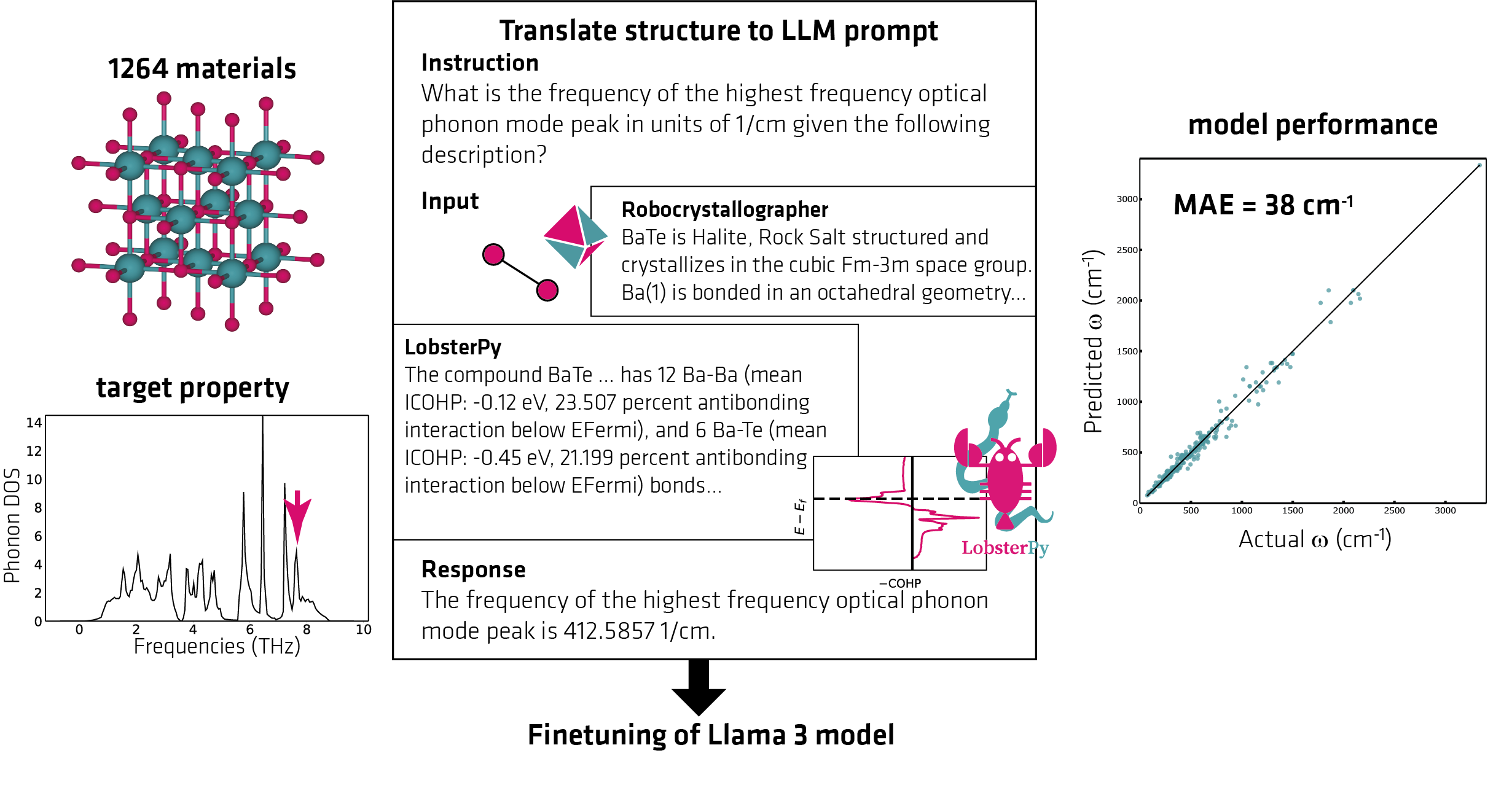}
    \caption{Schematic depicting the prompt for fine-tuning the LLM with Alpaca prompt format.}
    \label{fig:lobster}
\end{figure}

During the hackathon, one Llama model was fine-tuned with the Alpaca prompt format using both Robocrystallographer and LobsterPy text descriptions, and another one using solely Robocrystallographer input. \autoref{fig:lobster} depicts the prompt used to fine-tune an LLM to predict the last phonon DOS peak. The train/test/validation split was 0.64/0.2/0.16. The models were trained for 10 epochs with a validation step after each epoch. The textual output was converted back into numerical frequency values for the computation of MAEs and RMSEs. The results show that including bonding-based information improved the model's prediction. The results also corroborate the team's previous finding that quantum-chemical bond strengths are relevant for this particular target property. Both model performances (Robocrystallographer: 44 cm$^{-1}$, Robocrystallographer+LobsterPy: 38 cm$^{-1}$) are comparable to other models of the MatBench test suite, with MAEs ranging from 29 cm$^{-1}$ to 68 cm$^{-1}$ as per the time of writing~\cite{matbench_phonon_2024}. 


Although the preliminary results seem promising, the models have not yet been exhaustively analyzed, validated, or optimized yet. As the prediction of a numerical value and not its text embedding is of interest to the task, further model adaptation might be beneficial. For example, Rubungo et al.~\cite{rubungo2023llm} modified T5, an encoder-decoder model, for regression tasks by removing its decoder and adding a linear layer on top of its encoder. Halving the number of model parameters allowed them to fine-tune on longer input sequences, improving model performance. A recently published benchmark for LLMs in materials property prediction also suggests that fine-tuning models with fewer parameters improves the prediction of materials properties~\cite{rubungo2024llm4mat}.

With the available easy-to-use packages like Unsloth,~\cite{unsloth_2024} the team was able to integrate their materials data into fine-tuning an LLM for property prediction with very limited resources and time. Since these initial results, the work has been extended to a dataset of bonding-based text descriptions including \~13,000 crystalline materials.  In the future, the team aims to (1) test these text descriptions further to learn other thermal and elastic material properties like elastic constants and lattice thermal conductivity and (2) to extend further the text descriptions generated with the LobsterPy package to include, e.g., information on computed charges.


\section{Molecular and Material Design}

LLMs have also been applied to molecular and material design,
proving capable in both settings~\cite{Bhattacharya2024Large, Liu2024Multimodal, Jia2024LLMatDesign, Jang2024Can, Lu2024Generative}, especially if pre-trained or fine-tuned with domain-specific data~\cite{pmlr-v235-kristiadi24a}. However, despite these advancements, LLMs still face limitations in practical applications~\cite{Miret2024Are}.

\subsection{{Leveraging AI Agents for Designing Low Band Gap Metal-Organic Frameworks}}

\begin{figure}
    \centering
    \includegraphics[width=0.7\linewidth]{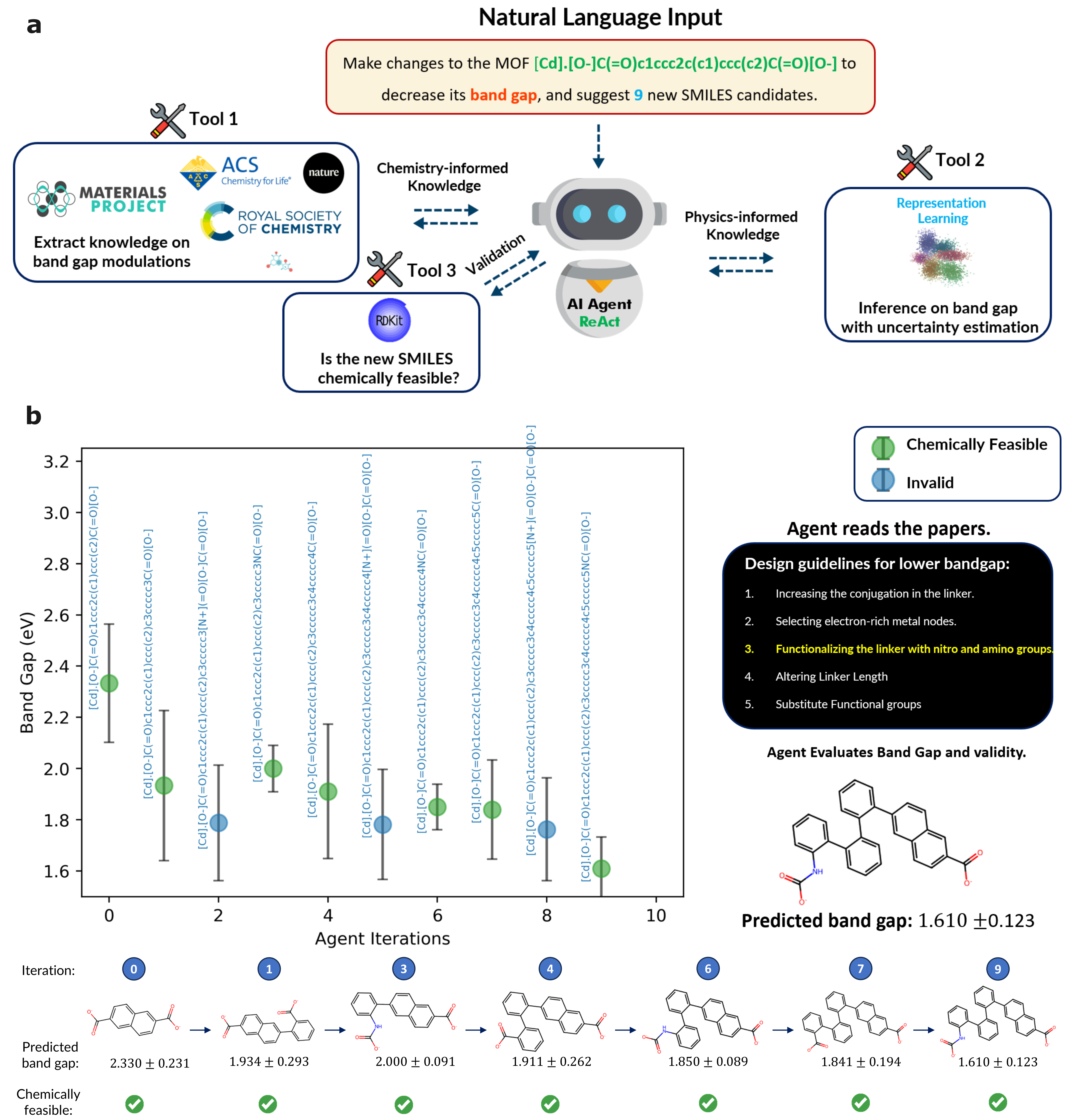}
    \caption{Workflow overview. The ReAct agent looks up guidelines for designing low band gap MOFs from research papers and suggests a new MOF (likely with a lower band gap). It then checks the validity of the new SMILES candidate and predicts the band gap with epistemic uncertainty estimation using an ensemble of surrogate fine-tuned MOFormers. b. Band gap predictions for new MOF candidates as a function of agent iterations}
    \label{fig:fig_section2}
\end{figure}

Metal-organic frameworks (MOFs) are known to be excellent candidates for electrocatalysis due to their large surface area, high adsorption capacity at low CO$_2$ concentrations, and the ability to fine-tune the spatial arrangement of active sites within their crystalline structure~\cite{li2022review}.
Low band gap MOFs are crucial as they efficiently absorb visible light and exhibit higher electrical conductivity, making them suitable for photocatalysis, solar energy conversion, sensors, and optoelectronics.
This submission aims at using chemistry-informed ReAct~\cite{yao2022react} AI Agents to optimize the band gap property of MOFs. The overview of the workflow is presented \ref{fig:fig_section2}a. 
The agent takes as inputs a textual representation of the initial MOF structure as a SMILES (Simplified Molecular Input Line-Entry System) string representation, and a short description of the property optimization task (i.e., reducing band gap), all in natural language.
This is followed by an iterative closed-loop suggestion of new MOF candidates with a lower band gap with uncertainty quantification, by adjusting the initial MOF given a set of design guidelines automatically obtained from the scientific literature.
A detailed analysis of this methodology, including its application to various classes of materials such as surfactants, ligands, and peptides can be found in reference~\cite{ansari2024dziner}, which supports both closed-loop and human-in-the-loop feedback cycles and thus enables real-time property inference for human-AI collaboration in molecular design.

The agent, powered by an LLM, is augmented with a set of tools allowing for chemistry-informed decision-making. 
These tools are as follows:
\begin{enumerate}
    \item \textbf{Retrieval-Augmented Generation (RAG)}:
    This tool allows the agent to obtain design guidelines on how to adapt the MOF structure from unstructured text. 
    Specifically, in this prototype, the agent has access to a fixed set of 7 MOF research papers (see Refs.~\cite{usman2017semiconductor, flage2013band, yang2014band, yang2010theoretical, ali2021recent, yan2023tuning, lin2012tunability}) as PDFs.
    This tool is designed to extract the most relevant sentences from papers in response to a given query. 
    It works by embedding both the paper and the query into numerical vectors using OpenAI's text-ada-002~\cite{greene2022new}, then identifying the top $k$ passages within the document that either explicitly mention or implicitly suggest the adaptations required for the specified band gap property for a MOF. 
    Inspired by the team's earlier work~\cite{ansari2024agent}, $k$ is set to 9, but is dynamically adjusted based on the relevant context's length to avoid OpenAI's token limitation. 
    
    \item \textbf{Surrogate Band Gap Predictor}
    The surrogate model used is a transformer (MOFormer~\cite{cao2023moformer}) that takes as input the MOF as a SMILES string. 
    This model is pre-trained using a self-supervised learning technique known as Barlow-Twin~\cite{zbontar2021barlow}, where representation learning is done against structure-based embeddings from a crystal graph convolutional neural network (CGCNN)~\cite{xie2018crystal}. This was done against 16,000 BW20K entries~\cite{moosavi2020understanding}. 
    The pre-trained weights are then transferred and fine-tuned to predict the band gap labels taken from 7,450 entries from the QMOF database~\cite{rosen2021machine}. 
    From a 5-fold training, an ensemble of five transformers are trained to return the mean band gap and the standard deviation, which is used to assess uncertainty for predictions. 
    For comparison, the team's transformer's mean absolute error (MAE) is approximately 0.467, whereas MOFormer, which was pre-trained on 400,000 entries, achieves an MAE of approximately 0.387.
        \item \textbf{Chemical Feasibility Evaluator} This tool primarily uses RDKit~\cite{landrum2013rdkit} to convert a SMILES string into an RDKit \emph{Mol} object, and performs several validation steps to ensure chemical feasibility. First, it parses the SMILES string to confirm correct syntax. 
        Next, it validates the atoms and bonds, ensuring they are chemically valid and recognized. 
        It then checks atomic valences to ensure each atom forms a reasonable number of bonds. 
        For ring structures, RDKit verifies the correct ring closure notation. Additionally, it adds implicit hydrogens to satisfy valence requirements and detects aromatic systems, marking relevant atoms and bonds as aromatic.
        These steps collectively ensure the molecule's basic chemical validity.
\end{enumerate}
The team has used OpenAI’s GPT-4~\cite{openai2023gpt4} with a temperature of 0.1 as the preferred LLM and LangChain~\cite{chase2022langchain} for the application framework development (nonetheless, the team confirms that the choice of LLM is only a hyperparameter and other LLMs can drive the agent).

The new MOF candidates and their corresponding inferred band gap are represented in ~\autoref{fig:fig_section2}b.
The agent starts by retrieving the following design guidelines for low band gap MOFs from research papers: 1. Increasing the conjugation in the linker. 
2. Selecting electron-rich metal nodes.
3. Functionalizing the linker with nitro and amino groups. 
4. Altering linker length.
5. Substitute functional groups (i.e., substituting hydrogen with electron-donating groups on the organic linker).
Note that the metal node adaptations are restrained by simply changing the system input prompt.
The agent iteratively implements the above strategies, makes changes to the initial MOF, and suggests a new SMILES.
The new SMILES is validated using the Chemical Feasibility Evaluator tool, and if found invalid, the agent uses a self-correction feedback loop to suggest new candidates, accounting for the extracted design guidelines.
After each valid modification, the band gap of the new MOF is then assessed using the fine-tuned ensemble of surrogate MOFormers to ensure a lower band gap. 
The self-correction feedback loop also handles new MOFs with undesired higher band gaps with respect to the initial MOF, by reverting to the most recent valid MOF candidate with the lowest band gap identified throughout the iterations.

\section{Automation and Novel Interfaces}

LLMs are increasingly important to the modern scientific workflow, enabling the development of more intuitive interfaces for users dealing with complex digital tools. For example, platforms such as ChemCrow~\cite{Bran2023ChemCrow}, RestGPT~\cite{Song2023RestGPT}, and HoneyComb~\cite{Zhang2024HoneyComb} allow researchers to input commands in natural language to interact with and analyze complex software and databases. With LLMs, democratized access and dramatically simpler interfaces are possible for programs like specialized computational techniques or command-line interfaces that may previously have required deep expertise.   LLMs excel at autonomous planning and task execution in multistep scenarios ~\cite{Boiko2023Autonomous} by breaking complex processes into smaller actions, making experimental or computational workflows controllable by models with less need for direct oversight. Such behavior may include but is not limited to: simple interaction with laboratory robotic systems ~\cite{Darvish2024ORGANA, Tom2024Self}, where difficult scientific objectives can be converted into precise, callable commands: the basis of precision and consistency. The integration of LLMs and robotics promises to improve operational efficiency and enable new designs of experimental workflows with increased flexibility. 

\vspace{0.7ex}
\noindent 
\subsection{LangSim – Large Language Model Interface for Atomistic Simulation}

LLMs can augment scientists with their common workflows, dramatically simplifying the interactions across systems using natural language input to understand and implement the intent of the user. The LangSim project ~\cite{LangSim} prototyped an interface to showcase the ability of LLMs to autonomously start atomistic simulations to study material properties on an atomistic scale. This provides the LLM with a way to request and then use novel scientific data and insights that were previously not available in published databases. One might imagine, e.g., the on-the-fly calculation of defect properties, e.g., grain boundary segregation energies 
In addition, by integrating the LLM in the active learning cycle of an autonomous materials discovery loop, with the option to calculate different material properties and access existing databases, the LLM becomes an AI scientist on a quest to discover novel materials. In this project, straightforward atomistic simulation and agentic scientific reasoning were explored as a natural language interface to users without programming skills.

The LangSim project implements atomistic simulation agents based on both pyiron~\cite{pyiron-paper} and LangChain~\cite{Bagatur2024langchain}. LangChain enables the LLM to call any kind of Python function and include the output in the thought process of the next iteration. In the case of LangSim, these Python functions represent simulation workflows implemented in the pyiron ~\cite{pyiron-paper} workflow framework to calculate material properties with atomistic simulations. By restricting the LLM to pre-defined simulation workflows, the risk of hallucination is reduced compared to generative approaches, which request the LLM to define and generate the simulation workflow. 
Based on the MACE~\cite{Batatia2024} foundation model for atomistic simulation, LangSim was used to predict the binary concentration of solid solution alloy required to match a user-defined bulk modulus, demonstrating an inverse materials design approach to enable application-specific alloy design. 

\begin{figure}
    \centering
    \includegraphics[width=0.9\linewidth]{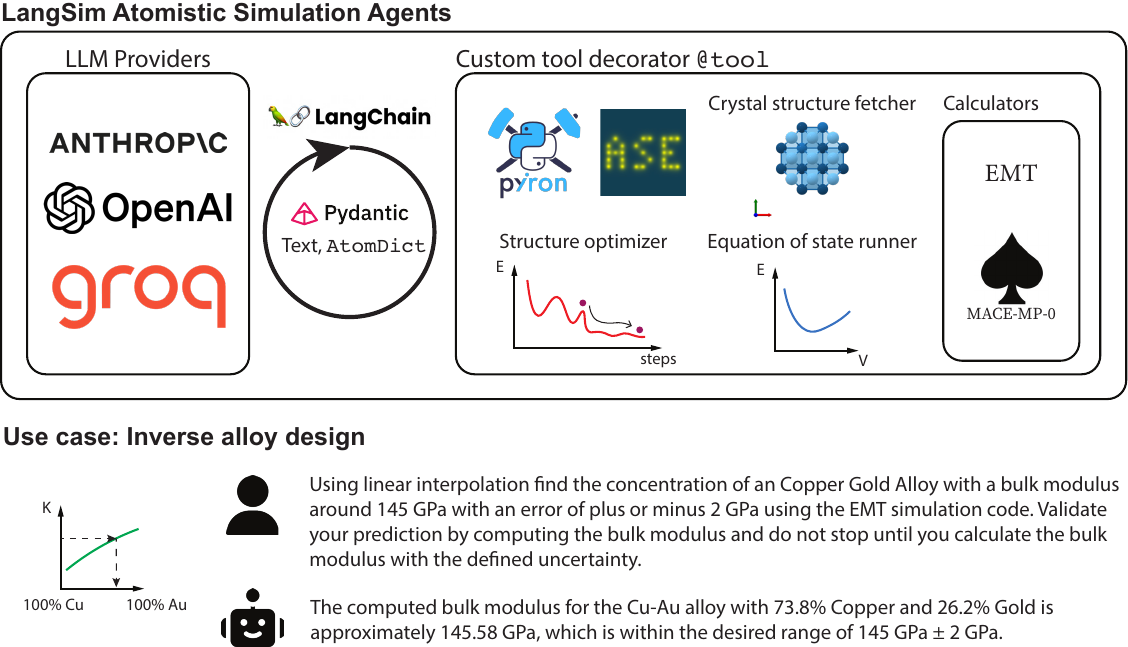}
    \caption{LangSim framework for atomistic simulation and inverse design. Custom atomistic modeling tools (such as \texttt{pyiron, ASE} python package functions with underlying EMT and MACE-MP-0 forcefields) are integrated using LangChain \texttt{@tool} decorator. Pydantic model is used to exchange atomic information in a structured format between LLM and tools. The emerging agentic capability for inverse alloy design is demonstrated. LLM agent is able to find the target composition of Cu-Au alloy with the desired bulk modulus.}
    \label{fig:langsim}
\end{figure}

\subsection{LLMicroscopilot: assisting microscope operations through LLMs}

While the state-of-the-art microscopes in materials science are crucial for high-resolution imaging and analysis, they are still rather addressed by expert operators due to their complex and steep-cost ownership. Their manipulation involves delicate tasks, mostly involving precision alignment, guaranteed optimal performances, and shifting between different operational modes to address various research questions that require extensive training and experience. This unobtainable quality has not only slowed down routine experimental procedures but has also formed a serious roadblock to opportunities for broadening access and allowing an acceleration of scientific discovery. With progress in natural language processing, LLMs opened the way for a Copernican revolution in this landscape. Integration of LLMs to the microscope interface will allow complex operations to be done through natural language commands. Similar to modern chatbots, which allow even those with no programming knowledge to generate complex computer programs ~\cite{Bauer2024}, LLMs can become intuitive intermediaries assisting users in traversing the manifold control procedures of advanced microscopes. Early studies of scanning probe microscopy have shown that LLMs can facilitate remote access ~\cite{Diao2024} and even direct control ~\cite{Liu2024} of these instruments, lessening the workload for expert operators. A promising approach is to use an LLM agent that accesses and operates some concrete external tools. These agents also interpret user commands and use observations in real time to make decisions, reducing the hallucinations, or wrong outputs, that sometimes appear with a standalone LLM. This would streamline the user experience, further relieving researchers from having to navigate through complex, tool-specific APIs, thus broadening the reach of advanced microscopes, especially to non-experts.

An illustration of such is the work performed by the LLMicroscopilot team, an LLM-based agent partially automating the operation of a scanning transmission electron microscope. LLMicroscopilot, its prototype, combines a generally trained foundation model, which is then tailored to specific people and domains through dedicated control tools. This agent operates quite well at first, utilizing the API for a microscope experiment simulation tool ~\cite{Madsen2021}, by performing such important tasks as estimating experimental parameters and executing the actual experiments. Therefore, this automation reduced dependence on personnel highly trained in operating such systems, thus increasing the opportunities for wider engagement in materials science due to the impact on usability. In the future, though, developments in the field are expected with LLMicroscopilot. In the future, they would involve integrating open-source microscope hardware control tools ~\cite{Meyer2019} and include capabilities for database access. Consequently, the system will be able to utilize Retrieval-Augmented Generation techniques to further inform parameter estimation and aid in the data analysis. Effectively, this will allow researchers to integrate LLMs in user interfaces at high-end microscopes and, instead of working on tedious, routine operational tasks, invest their energy in high-level scientific research and innovation, democratizing access to advanced experimental techniques.

\begin{figure}[h!]
    \centering
    \includegraphics[width=1\textwidth]{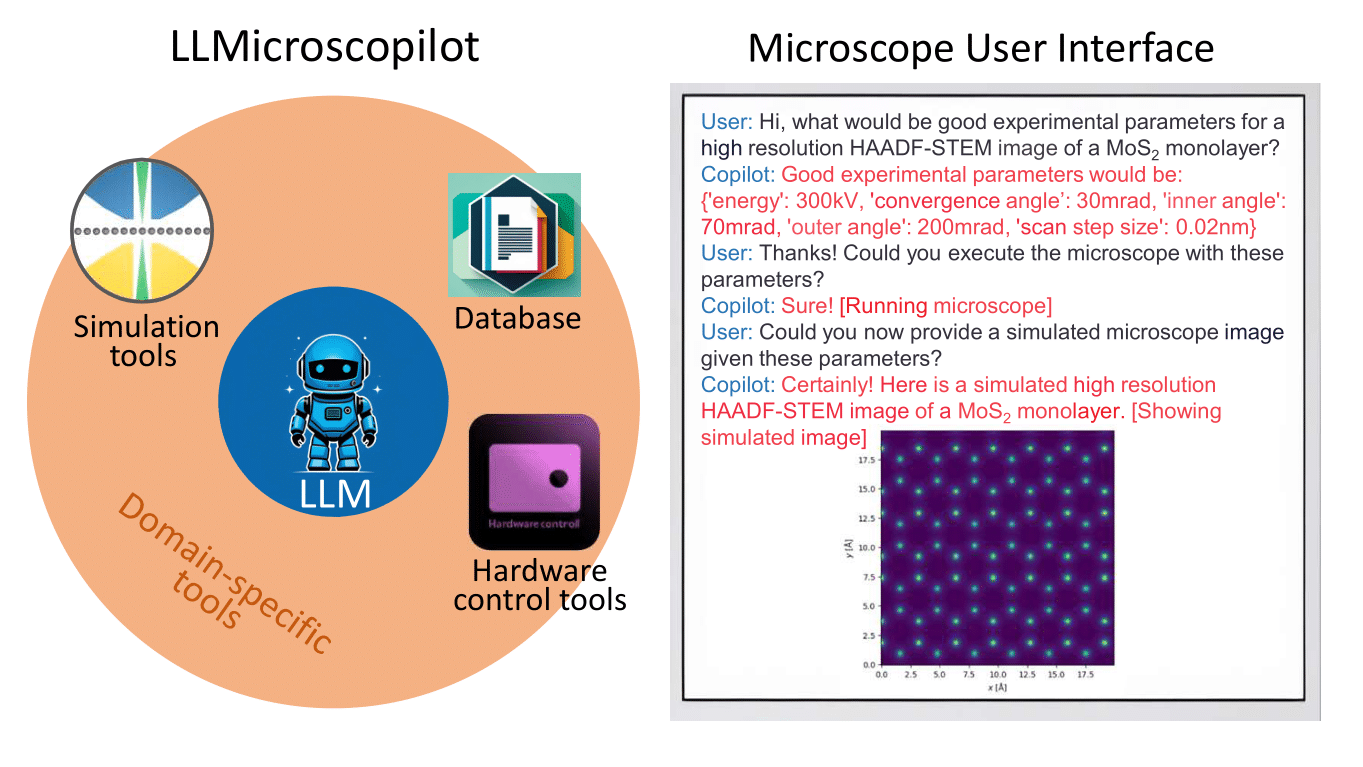} 
    \caption{Schematic overview of the LLMicroscopilot assistant. The microscope user interface allows the user to input queries, which are then processed by the LLM. The LLM executes appropriate tools to provide domain-specific knowledge, support data analysis, or operate the microscope.
    \label{fig:llmicroscopilot}}
\end{figure}

\section{Scientific Communication and Education}

LLMs are transforming how scientific and educational content is created and shared, enhancing accessibility and personalized learning~\cite{Yan2023Practical, Wang2024Large, Kasneci2023ChatGPT, Schäfer2023Notorious}. By automating tasks like question generation, feedback, and grading, LLMs streamline educational processes, freeing educators to focus on individual learning needs. Additionally, LLMs assist in translating complex scientific findings into accessible formats, broadening public engagement~\cite{Schäfer2023Notorious}. However, technological readiness, transparency, and ethical concerns around data privacy and bias remain critical challenges to address~\cite{Kasneci2023ChatGPT, Yan2023Practical}.

\subsection{MaSTeA: Materials Science Teaching Assistant}
This team selected 650 questions from the materials science question answering dataset (MaScQA)~\cite{zaki2024mascqa}, requiring undergraduate-level understanding to solve. These questions are classified into four types based on their structure: Multiple Choice Questions (MCQs), Match the Following (MATCH), Numerical Questions with Given Options (MCQN), and Numerical Questions (NUM). MCQs are generally conceptual, with four options, where mostly one is correct, though occasionally multiple answers are valid. MATCH questions involve two lists of entities that need to be correctly paired, with four answer choices provided, one of which contains the correct set of matches. MCQN questions present a numerical problem with four answer choices, requiring a solution to identify the correct option, while NUM questions have numerical answers rounded to the nearest integer or floating-point number as specified. The team aimed to automate the evaluation of open-source and proprietary LLMs on MaScQA and develop an interactive interface for students to engage with these questions. Various models, including LLAMA3-8B, HAIKU, SONNET, GPT-4, and OPUS, were evaluated across 14 subject categories, such as characterization, applications, properties, and behavior. The evaluation results, summarized in \autoref{evaluation on mascqa}, show that the OPUS variant of Claude consistently outperformed other models, achieving the highest accuracy in most categories. GPT-4 also demonstrated strong performance, particularly in material processing and fluid mechanics. As expected from prior studies, larger models such as OPUS and GPT-4 outperformed the smaller LLAMA3-8B, reinforcing the significance of model size in performance~\cite{sessler2024benchmarking}. The results suggest that there is significant room for improvement to enhance the accuracy of language models in answering scientific questions.

The evaluation involved:
\begin{itemize}
    \item \textbf{Extracting corresponding values:} For MCQs, correct choices were identified using regular expressions and compared to model predictions.
    \item \textbf{Prediction verification:} Numerical predictions were validated against exact or acceptable ranges, while MCQ responses were matched to correct answer choices.
    \item \textbf{Calculating accuracy:} Accuracy was computed per question type and topic, followed by an overall assessment across all questions.
\end{itemize}

\begin{figure}
    \centering
    \includegraphics[width=0.95\linewidth]{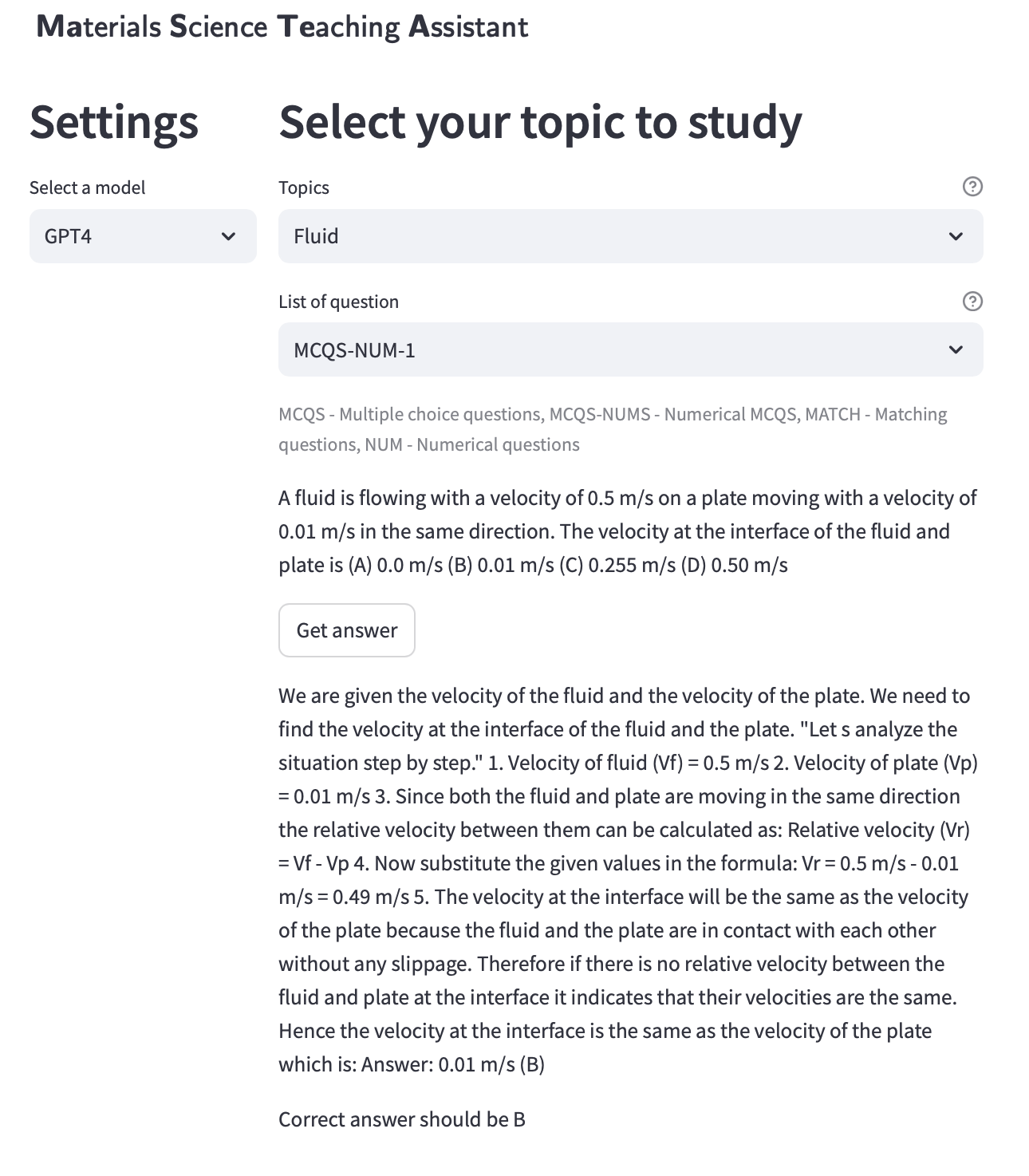}
    \caption{MaSTeA interface demonstrating a numerical question task. The model arrives at the correct answer by reasoning through the problem, providing students with a step-by-step solution if they struggle to solve it independently.}
    \label{fig:mastea}
\end{figure}

The evaluation results, summarized in \autoref{evaluation on mascqa}, show that the OPUS variant of Claude consistently outperformed other models, achieving the highest accuracy in most categories. GPT-4 also demonstrated strong performance, particularly in material processing and fluid mechanics. As expected from prior studies, larger models such as OPUS and GPT-4 outperformed the smaller LLAMA3-8B, reinforcing the significance of model size in performance~\cite{sessler2024benchmarking}. The results suggest that there is significant room for improvement to enhance the accuracy of language models in answering scientific questions. The interactive web app, MaSTeA (Materials Science Teaching Assistant), developed using Streamlit, allows easy model testing to identify LLMs' strengths and weaknesses in different materials science subfields. The interface can be seen in \autoref{fig:mastea}.


\begin{table}[ht!]
\centering
\caption{Accuracy of LLMs for each topic}
\scalebox{1.0}{
\begin{tabular}{lcccccc}
\hline
\textbf{Topic}                 & \textbf{\# Questions} & \textbf{LLaMA-3-8b} & \textbf{Haiku} & \textbf{Sonnet} & \textbf{OPUS} & \textbf{GPT4} \\ \hline
Thermodynamics        & 114   & 37.72  & 47.37  & 55.26  & \textbf{73.68}  & 57.02  \\ 
Atomic structure       & 100   & 32     & 40     & 49     & \textbf{64}     & 59     \\ 
Mechanical behavior    & 96    & 22.92  & 41.67  & 52.08  & \textbf{71.88}  & 43.75  \\ 
Material manufacturing & 91    & 43.96  & 57.14  & 56.04  & \textbf{80.22}  & 68.13  \\ 
Material applications  & 53    & 52.83  & 64.15  & 77.36  & \textbf{92.45}  & 86.79  \\ 
Phase transition       & 41    & 31.71  & 46.34  & \textbf{65.85}  & 70.73  & 63.41  \\ 
Electrical properties  & 36    & 33.33  & 25     & 55.56  & \textbf{72.22}  & 44.44  \\ 
Material processing    & 35    & 48.57  & 54.29  & 74.29  & \textbf{88.57}  & \textbf{88.57}  \\
Transport phenomena    & 24    & 37.5   & \textbf{70.83}  & 58.33  & 87.5   & 62.5   \\
Magnetic properties    & 15    & 26.67  & 46.67  & 46.67  & \textbf{66.67}  & 60     \\ 
Material characterization & 14    & 78.57  & 57.14  & 85.71  & \textbf{92.86}  & 71.43  \\ 
Fluid mechanics        & 14    & 21.43  & 50     & 57.14  & 78.57  & \textbf{85.71}  \\ 
Material testing       & 9     & 77.78  & 66.67  & \textbf{100}    & \textbf{100}    & \textbf{100}    \\
Miscellaneous          & 8     & 62.5   & 62.5   & 62.5   & \textbf{75}     & 62.5   \\ \hline
\end{tabular}
\label{evaluation on mascqa}
}
\end{table}

With MaSTeA, the team demonstrated the potential of interactive tools to help students practice answering questions and learn the steps to reach the correct solution. By evaluating LLM performance, the goal was to guide future model development and identify areas for improvement. The results suggest that LLMs can benefit from strategies such as self-consistency~\cite{wang2022selfconsistency} and retrieval-augmented generation (RAG)~\cite{lewis2020rag}, which have been shown to reduce hallucinations and increase accuracy. Additionally, integrating advanced reasoning models could further improve performance. Recent advancements in domain-specific LLMs, such as LLaMat~\cite{mishra2024foundational}, highlight the potential of specialized training to enhance scientific reasoning.

\section{Research Data Management and Automation}
Various submissions were received that attempt to enhance the management, accessibility, and automation of scientific data workflows using LLMs. These efforts, often leveraging multimodal agents, aim to simplify complex data handling, improve reproducibility, and accelerate insights across diverse scientific disciplines. We highlight two exemplar projects: ``yeLLowhaMmer'' a multimodal LLM-based data management agent that automates data handling within electronic lab notebooks (ELNs) and laboratory information management systems (LIMS), and ``NOMAD Query Reporter'', an LLM-based agent that uses RAG to generate context-aware summaries from large materials science repositories like NOMAD~\cite{Draxl2019NOMAD}
\subsection{yeLLowhaMMer: A Multi-modal Tool-calling Agent for Accelerated Research Data Management}
As scientific data continues to grow in volume and complexity, there is a need for tools that can simplify the job of managing this data to draw insights, increase reproducibility, and accelerate discovery. 
Digital systems of record, such as electronic lab notebooks (ELNs) or laboratory information management systems (LIMS), have been a great advancement in this area. 
However, capturing data using, e.g., electronic lab notebooks (ELNs) or laboratory information management systems (LIMS) is laborious, or simply impossible, to accomplish using graphical user interfaces alone. Recent advances in AI present an opportunity to augment how researchers interact with their data, improving scientific data management and allowing scientists to ask scientific questions of these data sources in new ways.

YeLLowhaMmer explored how large language models can be used to simplify and accelerate data handling tasks in order to generate new insights, improve reproducibility, and save time for researchers using the  open-source \emph{datalab}~\cite{Evans2024a} ELN/LIMS. Previously, the team had made progress toward this goal by developing a conversational assistant, Whinchat~\cite{hackathon-2023}, that allows users to ask questions about their data. 
However, this assistant was unable to take action with a user’s data or seek additional information as is often needed for scientific tasks. Thus, the team developed yeLLowhaMmer as a multimodal large language model (MLLM)-based data management agent capable of taking free-form text and image instructions from users and executing a variety of complex scientific data management tasks.

The agent is powered by commercial MLLMs used within an agentic framework capable of iteratively writing and executing Python code that interacts with \emph{datalab} instances via the \texttt{datalab-api} package. In typical usage, a yeLLowhaMmer user might instruct the agent: “Pull up my 10 most recent sample entries and summarize the synthetic approaches used.” In this case, the agent will attempt to write \emph{datalab} python API code to query for the user’s samples in the \emph{datalab} instance and write a human-readable summary based on the result. If the code it generates gives an error (or does not give sufficient information), the agent can iteratively rewrite the program until the task is accomplished successfully. Importantly, this paradigm is enabled by the presence of a structured API for diverse forms of scientific data; which is provided by \emph{datalab} in its open-source schemas and API documentation.

\begin{figure}[h!]
	\centering
	\includegraphics[width=0.7\textwidth]{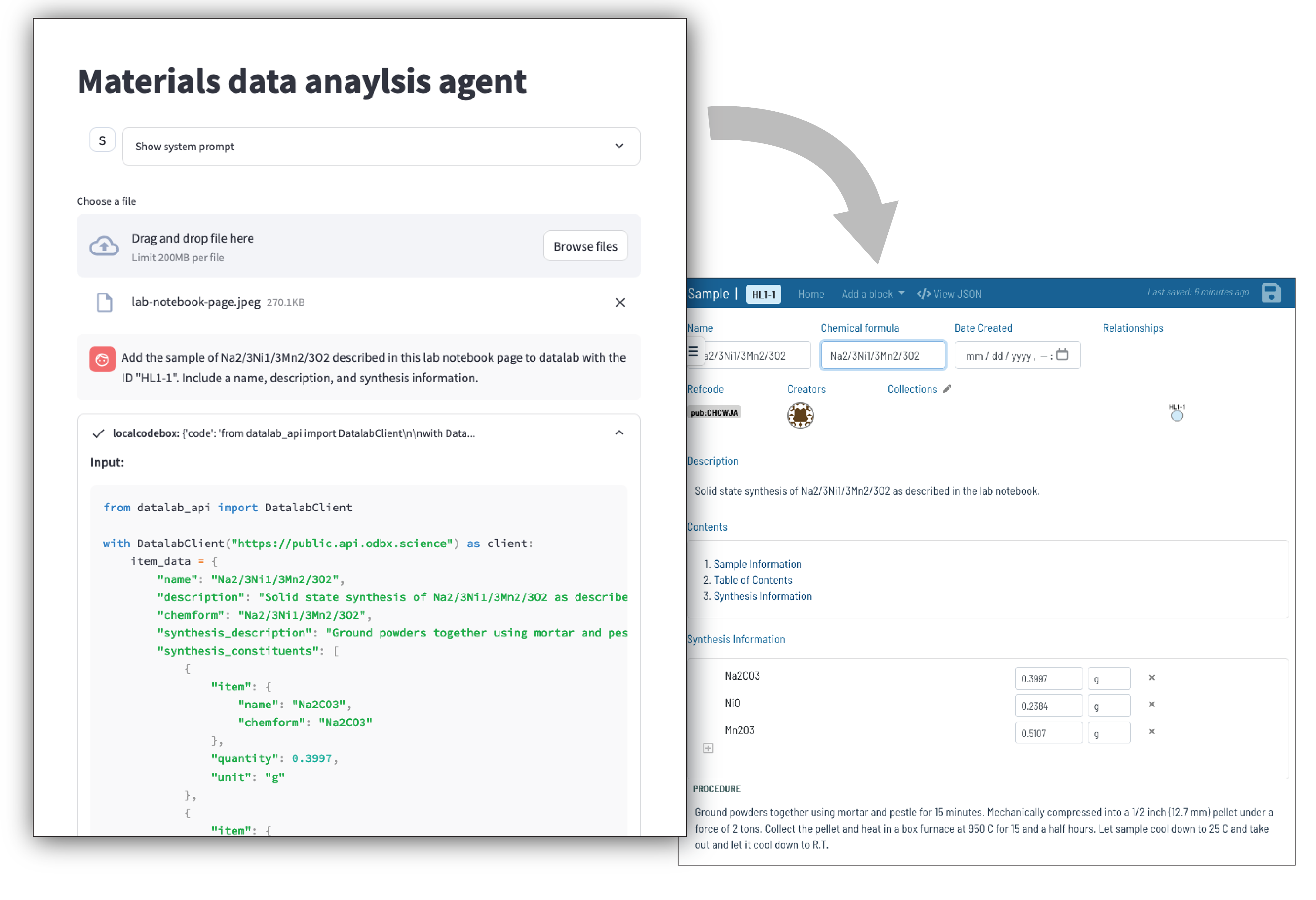} 
	\caption{The yeLLowhaMmer multimodal agent can be used for a variety of data management tasks. Here, it is shown automatically adding an entry into the \emph{datalab} lab data management system based on an image of a handwritten lab notebook page.}
	\label{fig:yellowhammer}
\end{figure}

In developing yeLLowhaMmer, the team found that simply copying documentation for the new datalab-api package into the system prompt produced poor code. Creating a simplified version with concrete examples and abridged JSON Schema formats proved more effective. The 12,000-character prompt (ca. 3,200 tokens) works well with modern large context models like Claude 3 Haiku. Future scientific libraries might benefit from maintaining both standard documentation and condensed "agents.txt" files optimized for ML agents.

This work shows the opportunity to integrate more tightly into scientific data management workflows to allow researchers to quickly handle complex tasks and efficiently ask questions of all collected  data. An important challenge is to find ways to ensure that data curated or modified by such agents will be appropriately `credited' by, for example, visually demarcating AI-generated content, and providing UI pathways for human users to verify or relabel such data in an efficient manner. Finally, recent progress in MLLM's ability to handle audio and video content in addition to text and images will allow agents to use audiovisual data in real time to provide even more comprehensive user interfaces. 

\subsection{NOMAD Query Reporter: Automating Research Data Narratives}

Research data management (RDM) in materials science includes a wide variety of schemas and data structures. Databases such as NOMAD~\cite{Draxl2019NOMAD, scheidgen2023nomad} support extensible context-aware schemas. Hence, the results of a single query may in fact contain various schemas, complicating the data analysis process.  NOMAD Query Reporter is a proof-of-concept application built to produce a written summary of the common methodological parameters and standout results in a scientific style. These may serve as the first step in an analysis workflow, or as progenitors of a journal article’s ``methods'' section.

\begin{figure}[h!]
	\centering
	\includegraphics[width=0.7\textwidth]{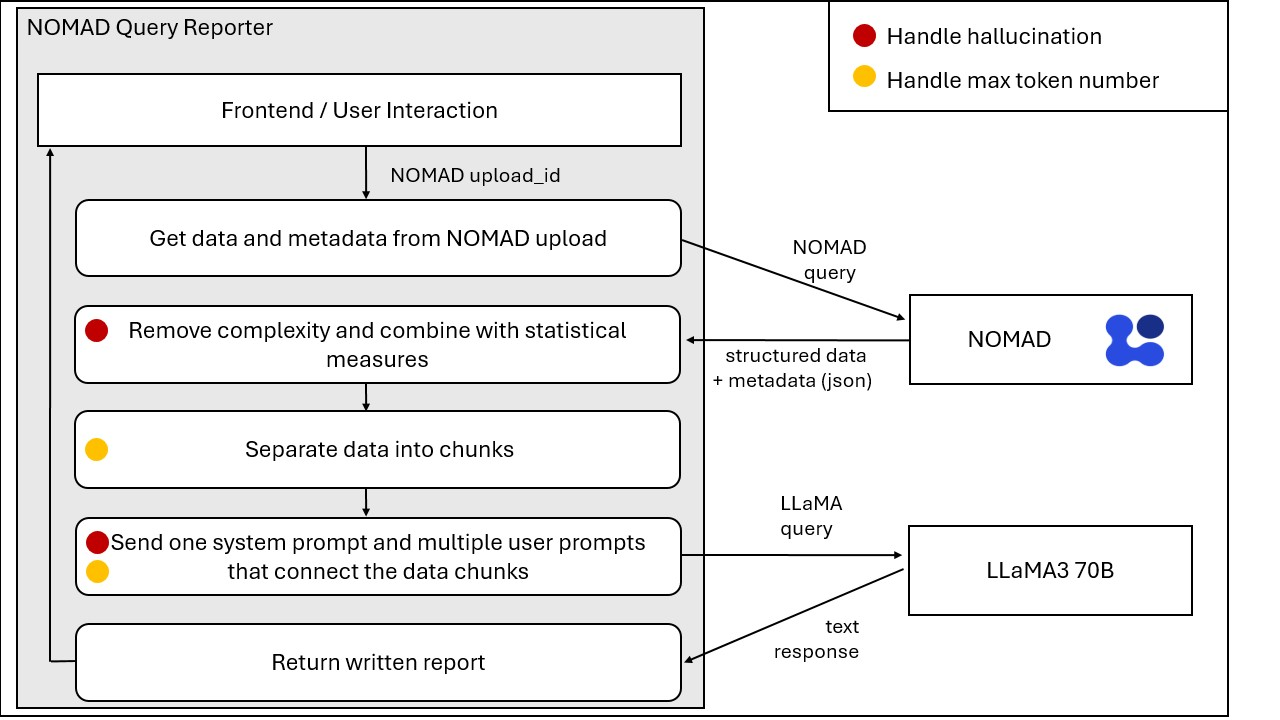} 
	\caption{Flowchart of the Query Reporter usage, including the back-end interaction with external resources,i.e., NOMAD and Llama. Intermediate steps managing hallucinations or token limits are marked in red and
orange, respectively.}
	\label{fig:nomad-query-reporter}
\end{figure}

Given the large size –over 19 million entries– and dynamic nature –open public uploads– of the NOMAD database, retraining or fine-tuning strategies are challenging. Instead, this prototype implements a retrieval-augmented generation (RAG) approach, as defined by Gao et al.~\cite{gao2023retrieval}, to enrich Llama3 (70B version) model's~\cite{touvron2023llama} knowledge base. The team progressively fed data by field into the LLM's chat-completion API as context. Subsequently,  the construction of the summary was completed by topic (i.e., properties, techniques, material composition) in a multi-turn conversation style with the ``roles'' feature clearly distinguishing the LLM's tasks from the data provided. Alignment with earlier versions of the chat history is enforced both via low-temperature settings as well as prompt engineering. For a step-by-step overview, see \autoref{fig:nomad-query-reporter}.

This work highlights the ability of LLMs to augment research data management systems via returning information in formats that are easily understandable by users. While the prototype NOMAD Query Reporter as able to manage homogenized hits well, attempts at extending to manually annotated,
heterogeneous data from ELNs proved challenging. Thus, follow-up work should consider more performant models and advanced RAG and other strategies to improve model context.

\section{Hypothesis Generation and Evaluation}
LLMs can be leveraged to streamline scientific inquiry, hypothesis generation, and verification. Recent work across psychology, astronomy, and biomedical research demonstrates their capacity to generate novel, validated hypotheses by integrating domain-specific data structures like causal graphs~\cite{Zhou2024Hypothesis, Abdel2024Scientific, Tong2024Automating, Ciuca2023Harnessing, Bazgir2025}. Although still largely untapped in chemistry and materials science, this approach holds substantial promise for accelerating discovery and innovation in these fields~\cite{Liu2024Beyond, Shir2024Towards, BazgirAdib2025}.

\vspace{0.7ex}
\noindent 
\subsection{Multi-Agent Hypothesis Generation and Verification through Tree of Thoughts and Retrieval Augmented Generation}
Scientific discovery thrives on the ability to generate and evaluate new hypotheses efficiently. However, the process of forming meaningful and testable hypotheses often requires extensive background research, domain knowledge, and iterative refinement. Advances in large language models offer an opportunity to assist researchers in streamlining this process, particularly through structured, multi-agent frameworks that systematically generate, evaluate, and refine ideas.

\begin{figure}
    \centering
    \includegraphics[width=0.9\linewidth]{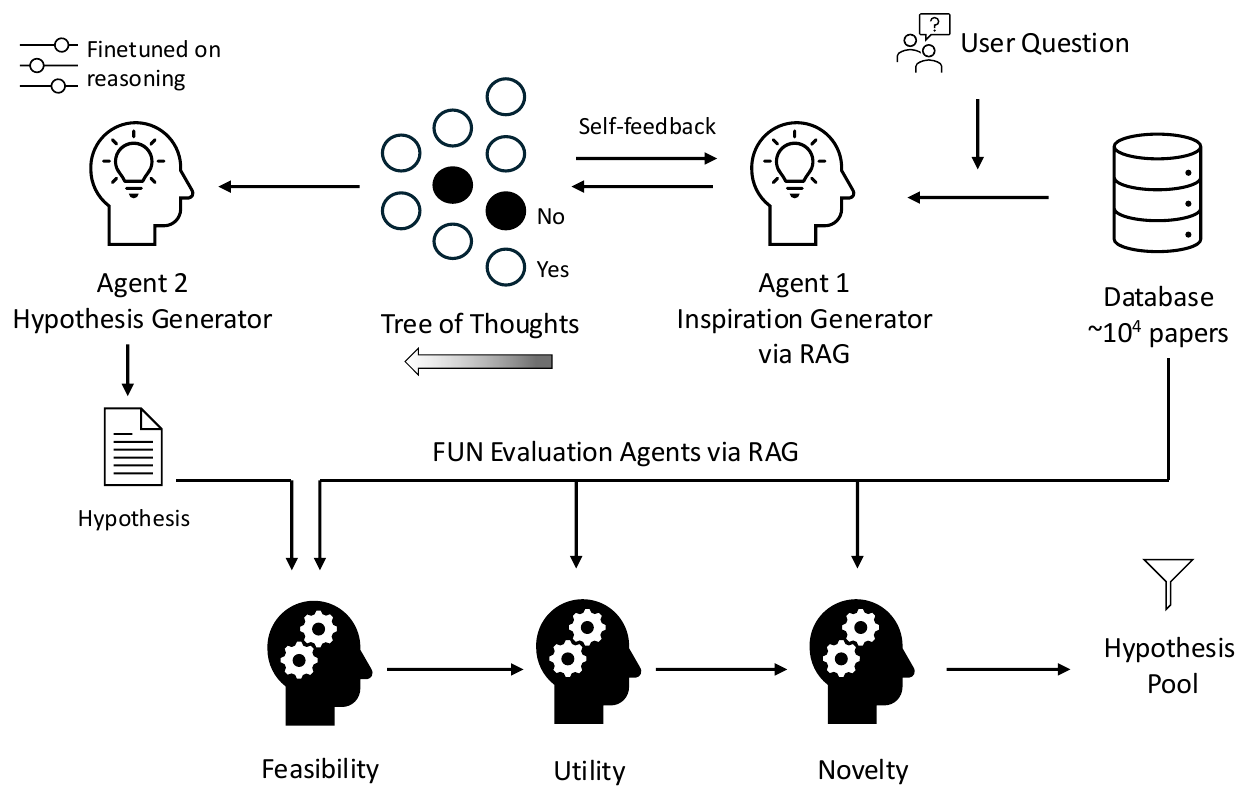}
    \caption{Multi-Agent Hypothesis Generation and Verification Framework. The system uses Retrieval-Augmented Generation, Tree of Thoughts, and Feasibility, Utility, and Novelty evaluation agents to generate and refine hypotheses for sustainable concrete design.}
    \label{fig:thoughtful_beavers}
\end{figure}

The Thoughtful Beavers team (Soroush Mahjoubi, Aleyna B.\ Ozhan) designed a multi-agent system to enhance scientific inquiry in materials science. Similar systems have proven useful in social sciences~\cite{yang2024largelanguagemodelsautomated}, and the system was adapted specifically for hypothesis generation in the domain of cement and concrete. The system consists of specialized agents that work in tandem: retrieving background knowledge, generating inspirations, formulating hypotheses, and evaluating their feasibility, utility, and novelty. By leveraging a combination of retrieval-augmented generation, tree-of-thoughts reasoning~\cite{NEURIPS2023_271db992}, and LLM-as-a-judge frameworks, this pipeline, which is illustrated in \autoref{fig:thoughtful_beavers}, ensures that only the most promising hypotheses emerge from the process.

To test this pipeline, the authors focused on sustainability challenges in concrete design. By processing 66,000 abstracts related to the field, an embedding-based retrieval system was built to extract relevant insights and generate research questions. From this dataset, the approach produced 1,000 structured hypotheses, which were then subjected to rigorous evaluation. The results showed that 243 hypotheses were deemed feasible based on current scientific knowledge, 175 demonstrated practical utility, and 12 stood out as highly novel. 

Looking ahead, this framework can be adapted to other material systems or even cross-disciplinary applications. By adjusting the background retrieval process, researchers could apply this method to areas such as ceramics, composites, or biomedical materials. Additionally, cross-pollination of ideas between domains could inspire new lines of research. As LLM capabilities continue to evolve, integrating AI-assisted hypothesis generation with expert validation could significantly accelerate scientific progress while maintaining the critical role of human creativity in innovation.

\section{Knowledge Extraction and Reasoning}
Extraction of structured scientific knowledge from unstructured text using LLMs to assisting researchers in navigating complex academic content is of wide interest~\cite{Shamsabadi2024Large, Dagdelen2024Structured, xu2024largelanguagemodelsgenerative, Li2024Generative}. These systems streamline tasks like named entity recognition and relation extraction, offering flexible solutions tailored to materials science and chemistry~\cite{Dagdelen2024Structured}. Tool-augmented frameworks help LLMs address complex reasoning by leveraging scientific tools and resources, expanding their utility as assistants in scientific research~\cite{ma2024sciagenttoolaugmentedlanguagemodels}.

\vspace{0.7ex}
\noindent
\subsection{ActiveScience}

\begin{figure}
    \centering
    \includegraphics[width=1\linewidth]{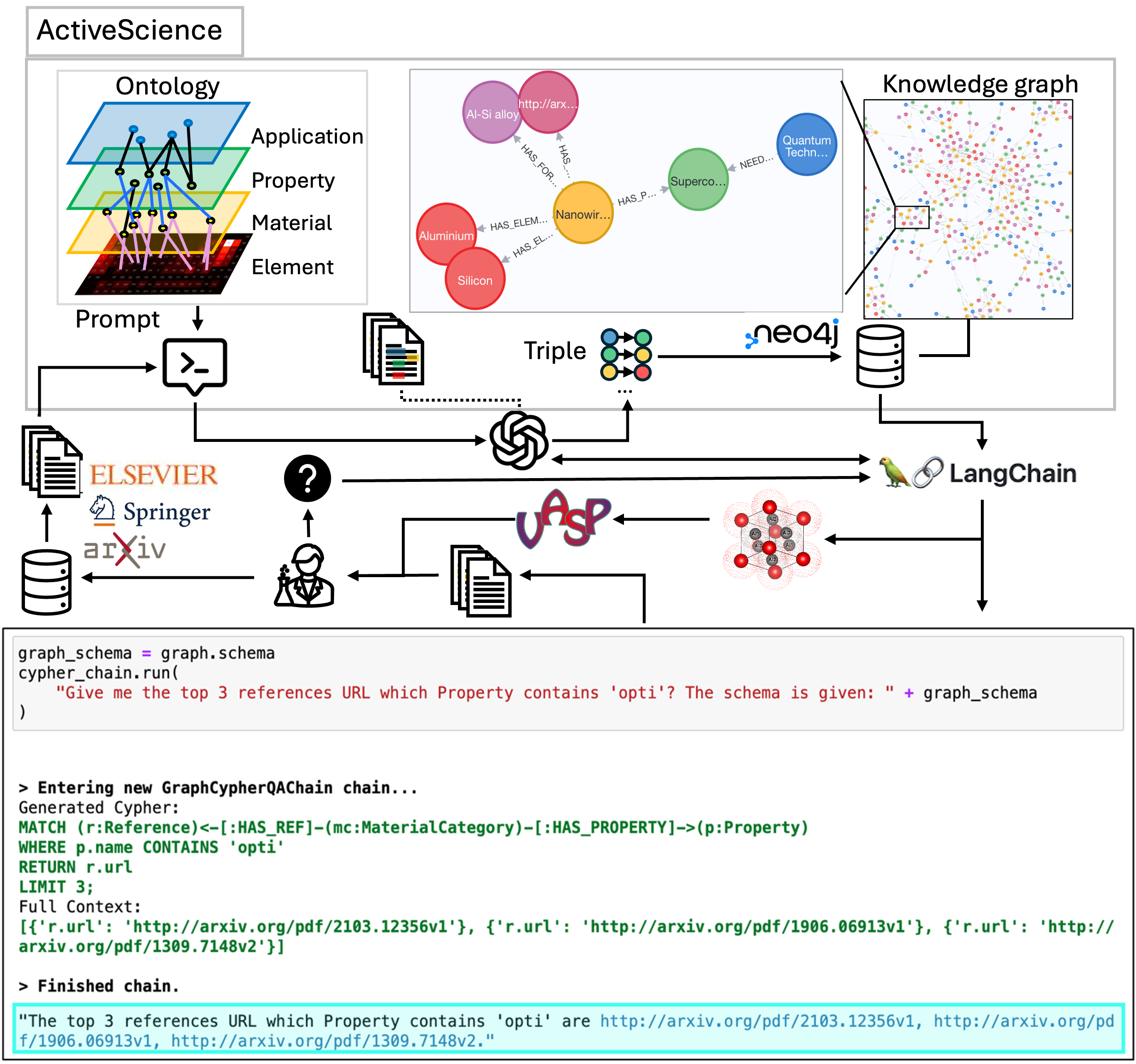}
    \caption{ActiveScience Framework for Knowledge Extraction. The system combines ontology-driven prompts, large language models, and a Neo4j knowledge graph to enable natural language queries and retrieval-augmented generation (RAG) for scientific research insights. Additionally, a code snippet demonstrating the use of LangChain is shown.}
    \label{fig:activescience}
\end{figure}

Extracting and refining knowledge in hard sciences is crucial. While large language models excel in summarization and dialogue generation, they are also prone to generating false information, a phenomenon known as hallucination. This presents a significant challenge for researchers leveraging LLMs in scientific fields. Various strategies exist to mitigate hallucinations. One approach involves fine-tuning models or constructing additional lightweight models after pretraining, but these methods require substantial computational resources, making them impractical in many cases. A more accessible alternative is retrieval-augmented generation (RAG), which enhances LLMs by incorporating external information. Conceptually, if a fine-tuned model resembles a domain expert with deep knowledge, a pre-trained model is akin to a generalist with broad understanding. By supplying additional context, pre-trained models can generate more accurate and reliable outputs.
To address this challenge, Min-Hsueh Chiu introduced an automated framework \emph{ActiveScience} that leverages large language models to ingest scientific articles into a knowledge graph and enable natural language queries for domain knowledge extraction. The framework integrates three key components: a data source API, a large language model, and a graph database. While these components can be replaced with equivalent technologies, this work specifically utilizes the ArXiv API~\cite{arxivapi}, GPT-3.5 Turbo~\cite{GPT35Turbo}, and Neo4j~\cite{Neo4J}.

For structured representation of knowledge and relationships, ActiveScience employs an ontology that defines key entities such as Application, Property, Material, Element, and Metadata. The ontology design is adaptable and scalable to specific use cases.
ActiveScience constructs its knowledge graph by extracting relevant triples from scientific articles. Specifically, prompts are generated using the predefined ontology and the introduction sections of articles to produce Cypher import statements containing structured triples, such as \textsf{(Material: "Nanowire") - [HAS\_ELEMENT] - (Element: "Aluminum")} and \textsf{(Material: "Nanowire") - [HAS\_FORMULA] - (Formula: "Al-Si alloy")}. These triples are then imported into a Neo4j graph database.
To facilitate RAG, the GraphCypherQAChain module from LangChain is employed. For instance, given the query, \textsf{"Retrieve the top three reference URLs where the Property contains ‘opti’?"}, GraphCypherQAChain dynamically generates a Cypher query based on the predefined ontology schema, executes it within Neo4j, and returns the relevant results. The processes of query generation and natural language processing are handled by LLMs. The pipeline and output are illustrated in \autoref{fig:activescience}.

\subsection{GlossaGen}
\begin{figure}
    \centering
    \includegraphics[width=0.9\linewidth]{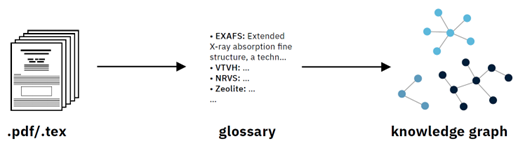}
    \caption{Schematic overview of the GlossaGen project. Textual information is extracted from PDF and LaTeX files and a glossary is generated with terms and their definition. From this, a knowledge graph is created, showing entities and relationships between terms.}
    \label{fig:glossagen}
\end{figure}

Academic literature, particularly review articles and grant applications, would substantially benefit from the inclusion of comprehensive glossaries elucidating complex terminology and discipline-specific nomenclature. However, the manual generation of such reference materials is a labor-intensive and redundant process. To address this limitation, Lederbauer et al. developed \emph{GlossaGen}, which leverages large language models to automate the creation of glossaries for academic articles and grant proposals, eliminating the need for time-consuming manual compilation. To efficiently process PDF or TeX articles, a pre-processing step automatically extracts the title and DOI, and chunks the text into smaller, context-preserving sections for LLM analysis. LLMs such as GPT-3.5-Turbo~\cite{GPT35Turbo} and GPT-4-Turbo~\cite{GPT4Turbo} then identify and define scientific terms with the help of Typed Predictors~\cite{khattab2023dspycompilingdeclarativelanguage} and Chain-of-Thought~\cite{wei2023chainofthoughtpromptingelicitsreasoning} prompting, ensuring well-structured, contextually relevant, and accurate outputs. The generated glossary is not merely presented as a list of terms but also as an ontology-based knowledge graph using Neo4J~\cite{Neo4J} and Graph Maker~\cite{GraphMaker}, visualizing the intricate relationships between various technical concepts (\autoref{fig:glossagen}). A user-friendly interface prototype, developed with Gradio~\cite{Gradio}, enables seamless interaction and customization, making the system accessible to researchers.

Future enhancements could focus on improving glossary output through LLM fine-tuning, integrating retrieval-augmented generation, and enabling article image parsing. Additionally, the system can better support users by allowing them to input specific terms for glossary explanations, ensuring comprehensive coverage even when LLMs omit key concepts. Overall, GlossaGen's rapid development and promising capabilities highlight the potential of LLMs to assist researchers in their scientific outreach.
\subsection{ChemQA}
Foundation models exhibit strong capabilities in chemistry reasoning, yet their performance across different input modalities — text, images, and their combination, remains underexplored. Building upon prior benchmarks such as IsoBench~\cite{fu2024isobenchbenchmarkingmultimodalfoundation} and ChemLLMBench~\cite{guo2023largelanguagemodelschemistry}, the \textbf{VizChem} team (Khalighinejad et al.) introduced \emph{ChemQA}\cite{chemQA2024}, a multimodal question-answering dataset designed to assess chemistry reasoning in language models.

\begin{figure}[h]
    \centering
    \includegraphics[scale=0.7]{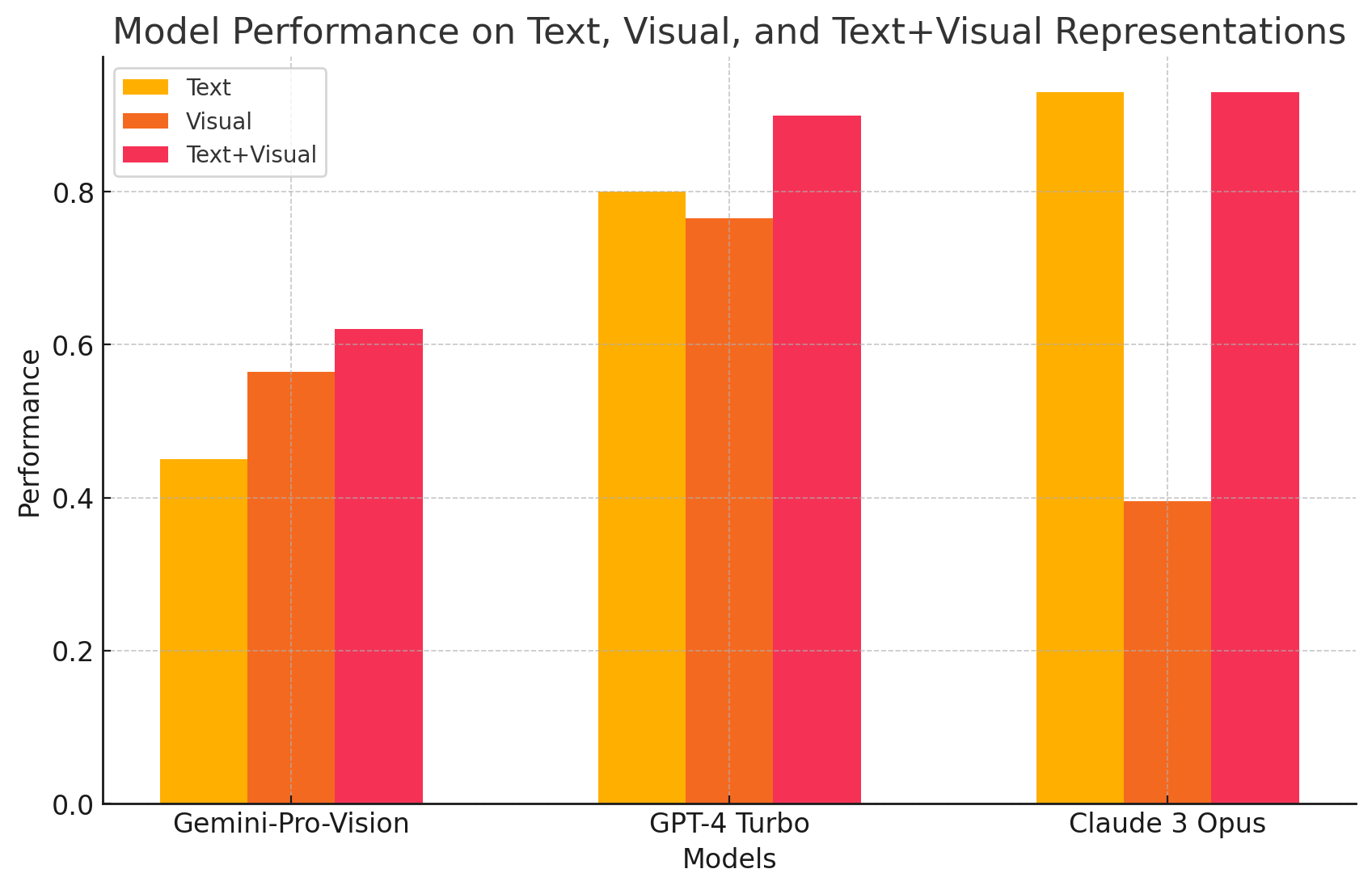}
    \caption{Performance of Gemini Pro, GPT-4 Turbo, and Claude3 Opus on text, visual, and text+visual representations. The plot shows that models achieve higher accuracy with combined text and visual inputs compared to visual-only inputs.}
    \label{fig:chemqa}
\end{figure}

ChemQA comprises five distinct QA tasks: atom counting, molecular weight calculation, name conversion, molecule captioning, and retrosynthesis planning. Each task is formulated with both molecular images and textual SMILES representations, enabling a systematic study of multimodal reasoning in chemistry.

The evaluation results, shown in \autoref{fig:chemqa}, reveal that the models achieve higher accuracy when provided with both text and images, while the performance drops significantly with image-only inputs. Notably, Claude 3 Opus demonstrates superior performance in text-based tasks, whereas Gemini Pro and GPT-4 Turbo excel in multimodal settings~\cite{openai2024gpt4technicalreport, geminiteam2024geminifamilyhighlycapable, TheC3}. These findings highlight the limitations of current models in processing visual chemistry data independently.

By introducing ChemQA, the VizChem team underscored the need for enhanced multimodal reasoning in chemistry. Future work should focus on improving the integration of textual and visual representations to advance AI-driven scientific analysis.

\section*{Hackathon Event Overview}
\begin{figure}
    \centering
    \includegraphics[width=0.75\linewidth]{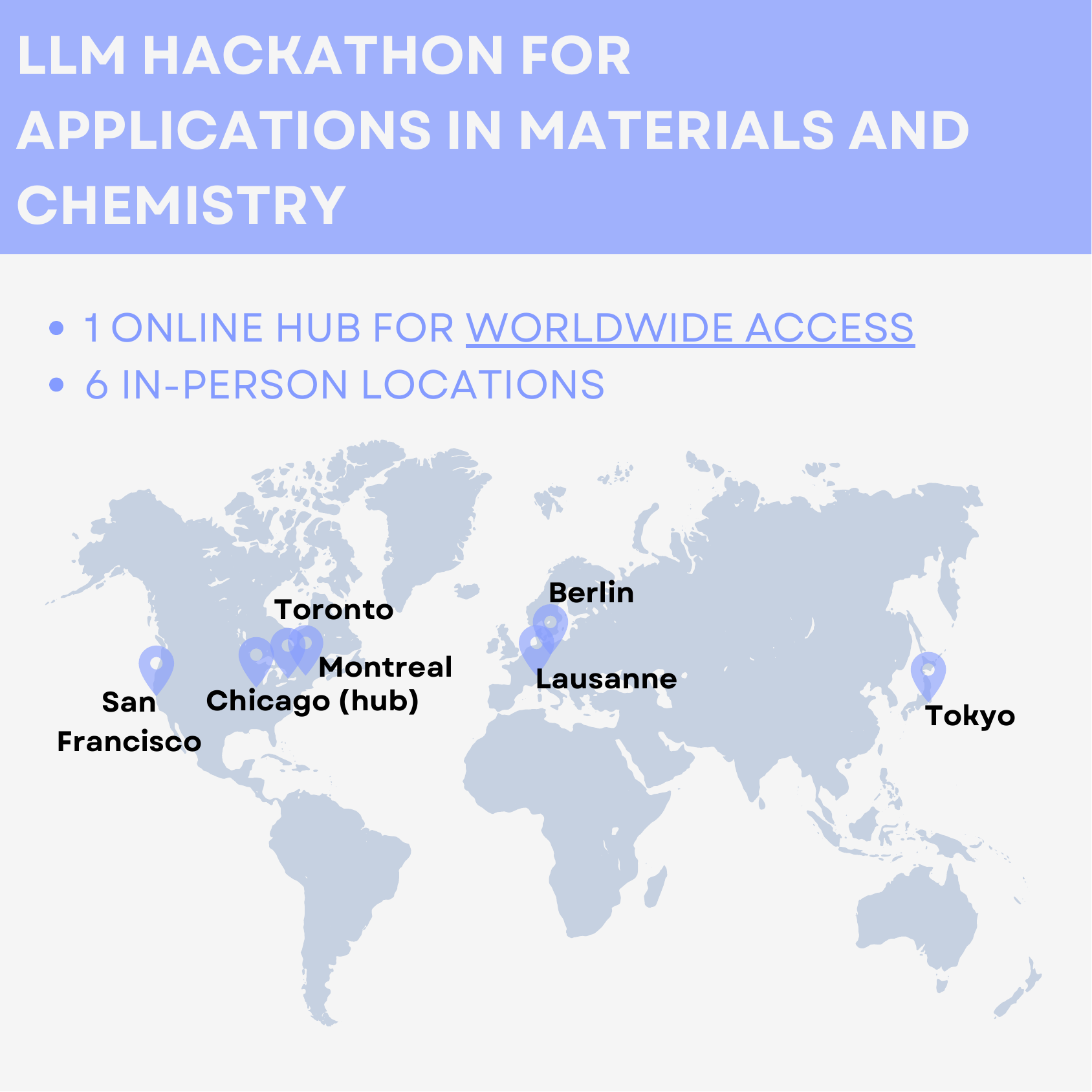}
    \caption{LLM Hackathon for Applications in Materials and Chemistry hybrid hackathon. Researchers were able to participate from both remote and in-person locations (purple pins).}
    \label{fig:map}
\end{figure}

The second annual Large Language Model (LLM) Hackathon for Applications in Materials Science and Chemistry was held on May 9, 2024, bringing together a global network of researchers, students, and industry professionals. With 556 registered participants and over 120 active contributors forming 34 teams, the event spanned multiple time zones and research domains, underscoring the broad interest in applying LLMs to scientific discovery(\autoref{fig:map}). This hackathon built on the success of the previous year’s event, described in detail in~\cite{hackathon-2023}.
The hybrid format included physical hubs in Toronto, Montreal, San Francisco, Berlin, Lausanne, and Tokyo, fostering interdisciplinary collaboration across institutions and time zones. The event began with a kickoff panel featuring experts Elsa Olivetti (MIT), Jon Reifsneider (Duke), Michael Craig (Valence Laboratories), and Marwin Segler (Microsoft), who discussed the evolving role of LLMs in scientific research.

The charge of the hackathon was intentionally open-ended: to explore the vast potential application space and create tangible demonstrations of the most innovative, impactful, and scalable solutions within a constrained timeframe. Participants leveraged open-source and best-in-class multimodal models to tackle challenges in materials science and chemistry. These teams submitted projects covering molecular property prediction, materials design, automation, hypothesis generation, and knowledge extraction, demonstrating the versatility of LLMs in scientific research. Many incorporated retrieval-augmented generation (RAG), multi-agent reasoning, and natural language interfaces, showcasing AI’s expanding role in scientific discovery.

Beyond technical contributions, the hackathon fostered a global research community, with 483 researchers continuing collaborations via Slack. The event demonstrated the value of structured collaboration in accelerating AI-driven discovery and bridging computational scientists, experimentalists, and AI researchers.


\section*{Conclusion}

The LLM Hackathon for Applications in Materials Science and Chemistry has demonstrated the dual utility and immense promise of LLMs to impact materials science and chemistry research across the entire lifecycle. Together, the projects  1) demonstrate the promise of a new set of tools that together form a cohesive patchwork to perform tasks ranging from hypothesis generation to data extraction, novel interface design, analysis of results, and more; and 2) showcase the ability of LLMs to enable rapid prototyping and exploration of the application space. Participants effectively utilized LLMs to explore solutions to specific challenges while rapidly evaluating their ideas over just a short 24-hour period, highlighting compelling abilities to enhance the efficiency and creativity of research processes across many applications. It's important to note that many projects benefited from significant advancements in LLM performance since the previous year's hackathon. That is, the performance across the application space was improved simply via the release of more powerful versions of Gemini, ChatGPT, Claude, Llama, and other models and more easily accessible APIs and examples. If this trend continues, we expect to see even broader applications in subsequent hackathons and in materials science and chemistry more generally. We note that reliance on proprietary APIs raises reproducibility concerns as models evolve or are deprecated, while infrastructure demands for training, fine-tuning, or running inference on models with parameters reaching hundreds of billions require yet more computational resources, leading to significant infrastructure roadblocks to further academic work.

Importantly, the hybrid hackathon format itself proved to be an effective mechanism to foster interdisciplinary collaboration, accelerate the prototyping of AI-driven tools, and create a global community of researchers engaged in exploring LLM applications. The hybrid format, combining physical hubs with virtual participation, facilitated knowledge exchange across continents, highlighting the importance of accessible, multimodal, and scalable approaches to scientific innovation.

\section*{Acknowledgments}
Planning for this event was supported by NSF Awards \#2226419 and \#2209892. We would like to thank event sponsors who provided platform credits and prizes for teams, including RadicalAI, Iteratec, Reincarnate, Acceleration Consortium, and Neo4j.
Site coordinators include: Brandon Lines, Philippe Schwaller, Pepe Marquez, 
Mehrad Ansari and Seyed Mohamad Moosavi. Mohamad Moosavi acknowledges support from the Data Science Institute at the University of Toronto for organizing events related to LLMs. Mehrad Ansari acknowledges  Mahyar Rajabi, Seyed Mohamad Moosavi, and Amro Aswad for their feedback on the project. Aakash Naik, Katharina Ueltzen, and Janine George would like to acknowledge the Gauss Centre for Supercomputing e.V. (www.gauss-centre.eu) for funding their work on this project by providing generous computing time on the GCS Supercomputer SuperMUC-NG at Leibniz Super computing Centre (www.lrz.de) (project pn73da).

\section*{Conflicts of Interest}
The authors declare no conflicts of interest. 

\FloatBarrier
\addcontentsline{toc}{chapter}{References}
\phantomsection
\bibliographystyle{ieeetr}
\bibliography{references}


\end{document}